\documentclass{article}
\usepackage{graphicx} %
\usepackage{amsmath}
\usepackage{amsthm}
\usepackage{subcaption}

\usepackage{xcolor}
\usepackage{array}
\usepackage{booktabs}
\usepackage{tabularx}
\usepackage{url}
\usepackage[T1]{fontenc}
\usepackage{authblk}
\usepackage{tikz}
\usetikzlibrary{arrows.meta, positioning, fit, calc}
\usepackage{natbib}
\usepackage{hyperref}

\title{Value Over Language Model: Detecting Original Contribution in Writing}

\author{Vibhhu Sharma}
\author{Thorsten Joachims}
\author{Sarah Dean}

\affil{Cornell University}

\date{August 2026}

\begin{document}

\maketitle

\begin{abstract}

Large Language Models have been rapidly adopted across writing tasks, prompting the development of tools for detecting LLM-generated text. However, these tools largely measure how much of a document's surface text was written by an LLM and aren't fundamentally designed to measure how much of the information content or ideas originated from the LLM itself rather than being supplied by the user in the prompt. In this work, we design a statistically backed framework that measures how much value a person adds on top of what a language model could have produced by itself. The method requires no training or labeled data and never scores the document's surface text, insulating it from stylistic confounders. Instead, it extracts the document's content at increasing levels of granularity, uses an LLM to reconstruct the document from each partial representation, and compares these reconstructions with those produced from the task description alone. We call this framework \textit{Value Over Language Model} (VOLM), which measures a document's contribution relative to a replacement-level document that an LLM could produce from the task description alone. We evaluate VOLM with a specific instantiation of this framework across three domains: news articles, ICLR peer reviews, and argumentative essays. VOLM separates human-ideated -- but potentially LLM-written -- documents from LLM-generated documents produced from generic task descriptions, while remaining substantially invariant to content-preserving transformations. We further find that increasingly constrained content extractors reduce residual differences between LLM-generated and humanized text, demonstrating the importance of disentangling informational content from stylistic variation. We hope these results encourage further work on specialized instantiations of the framework and on assessing human contributions in LLM-assisted writing more generally.

\end{abstract}

\section{Introduction}
\label{sec:intro}

Since their large-scale commercial introduction in late 2022, large language models have seen increasing adoption across writing tasks. 
An October 2025 study into AI use in American newspapers \citep{russell2026aiuseamericannewspapers} used a state-of-the-art LLM detector \citep{thai2025editlensquantifyingextentai} to find that approximately 9\% of newly-published articles in the summer of 2025 were either partially or fully AI-generated. Another recent survey \citep{Naddaf2025MoreTH} of 1600 academics reported that over 50\% of researchers have used artificial intelligence at some point while peer reviewing manuscripts, and 24\% of these researchers increased their use of AI for peer review in the past year. These statistics indicate that LLM use is widespread, but they reveal little about how these tools are actually used. The same reported ``use" can describe an author who hands a model a single-line prompt and submits whatever comes back, or one who provides a fully formed argument and asks the model only to help render it into prose. This distinction is important because it determines whose ideas the finished document actually expresses.

Consider a reviewer who reads a paper carefully and identifies specific flaws, for instance, ``\textit{the proof of Lemma 3 breaks down when n=2, the baseline in Table 2 is run with a different learning rate than the proposed method, the related-work section omits a closely related result from Smith et al. and the main theorem's assumption of bounded noise is not stated in the abstract.}" She then asks an LLM to expand this into a full review, complete with context, phrasing, and structure. Because the reviewer supplied only the underlying point and the LLM produced most of the surface text, a detector measuring the fraction of LLM-written text would likely flag this review as LLM-assisted. Yet, in this case, the reviewer's original contribution, i.e., identifying the flaw in the first place, is exactly the kind of substantive engagement peer review depends on, and the LLM's role was to help express it, not to originate it. Treating this review as suspicious simply because most of its words came from a model conflates the amount of LLM-produced text with the source of the review's content, and risks discouraging reviewers from making the best use of LLMs to help them communicate more effectively. This distinction is also important for researchers writing in a non-native language. For instance, a reviewer might formulate an argument in her native language and use an LLM to translate it into English. In this case, even though the model generated all of the document's English surface text, the argument and supporting ideas are still the author's own. A measure that equates extensive LLM-written text with limited human contribution would thus penalize exactly the kinds of translation and linguistic assistance that can make scholarly communication more accessible.

Existing work around LLM use in writing largely asks the binary question of whether a given piece of text was formulated by a language model or not \citep{liaodetection, liang2024mappingincreasingusellms, thai2025editlensquantifyingextentai}. The implicit aim is to identify machine-generated text so that it can be flagged, filtered, or penalized. A more recent line of work refines this by attempting to measure the fraction of text in a document that was subject to ``LLM editing'' instead, providing a more granular estimate of LLM involvement in writing in the form of a real number between 0 and 1 \citep{thai2025editlensquantifyingextentai}. However, both framings retain the same core motivation, as they propose to measure the amount of LLM involvement in writing the final text, instead of the more involved, and arguably more important, question of \textit{whose ideas the text expresses}. A document can be almost entirely rendered into prose by a model but still be the product of substantial original thought by a human author, if the author provided detailed information and guidance to the language model as part of their prompt. As such, we aim to design a methodology that definitionally uncovers the extent of original contribution that went into the creation of a document, instead of measuring a document's closeness to LLM-style writing.

We measure this extent as the gap between what the human supplied as part of their prompt and what a language model was already capable of producing on its own with a generic prompt. If a model can produce a document from a short, generic prompt alone, the content was already latent in its weights (or readily available to the model through tools), and the author's contribution was minimal. Meanwhile, if producing the same document requires extensive guidance, such as a detailed structure and specific arguments, then the human supplied information that the model did not have. 

We operationalize this idea by asking, "how much of a document’s content must be removed before a language model finds it no more surprising than a document generated from the task description alone?" Modern LLMs produce text autoregressively \citep{openai2024gpt4ocard}, assigning each token a probability conditioned on the tokens before it; the negative log of this probability is the token's surprisal, and a document's average surprisal under a model measures how well the model could have anticipated it. A document whose content is already latent in the model, recoverable from a short generic prompt, should be met with low surprisal. On the other hand, a document containing information the human author supplied and the model did not already have should be met with higher surprisal, and this surprisal should persist even as more of the document's own content is revealed to the model as context.

We name our framework Value Over Language Model (VOLM), borrowing both the name and the underlying premise from Value Over Replacement Player (VORP), an advanced statistic from baseball sabermetrics popularized by Keith Woolner. VORP does not ask how good a player is in some absolute sense, instead asking how much better a player is than a \textit{replacement-level} player, i.e., one performing at the level a team could obtain at minimal cost, from the pool of freely available talent on the waiver wire. 

We adapt this logic to writing. In this context, the role of a 'replacement-level player' is taken over by a language model, and we measure a human's contribution on top of what a language model could have easily written by itself with a generic task prompt. More simply, the baseline replacement-level document is the one a language model would produce given nothing but the generic task description. An author's value, like a player's, is not their output in isolation, but their output above what was already freely available.

Our contributions are as follows: 
First, we reframe the problem of measuring LLM involvement in writing. Instead of measuring how much of a document's text a model produced, we measure how much of a document's 'information' originated with its human author, a distinction current detectors do not draw. 
Second, we introduce VOLM, a general framework for quantifying this original contribution relative to a model, applicable to any document paired with a task description. 
Third, we develop a greedy algorithm that implements VOLM, producing both a document-level score and identifying the specific high-VOLM content in the document. This algorithm requires no model training and no labeled data, operating entirely through log-probabilities computed with an off-the-shelf language model. We also propose and test a computationally-cheaper alternative to score documents, finding that it achieves similar separation between human-ideated and LLM-generated documents at a lower computational cost.
Fourth, we show VOLM is robust to transformations of a document's surface text that leave its information content unchanged, including LLM-paraphrasing, translation round-trips, and, to a meaningful degree, humanizer tools explicitly designed to evade LLM detection, while remaining sensitive to genuine differences in original content. We evaluate VOLM across three domains: news articles, ICLR peer reviews, and argumentative essays.

\section{Related Work}

A large body of work attempts to determine whether a given document was produced by a language model. Zero-shot methods such as DetectGPT \citep{mitchell2023detectgptzeroshotmachinegeneratedtext} require no classifier training at all.
They observe that minor rewrites of model-generated text tend to have lower log probability under the model than the original sample, while minor rewrites of human-written text may have higher or lower log probability than the original sample. Meanwhile, \citet{liang2024mappingincreasingusellms, liaodetection} use a statistical approach to LLM detection by positing that human-written text and LLM-written text come from two separate distributions and explicitly defining these distributions for different domains. These distributions are modeled as token-specific probabilities and derived from real, labeled data. More recent work relaxes the binary framing and instead estimates the \textit{fraction} of a document attributable to LLM involvement, training a deep supervised model that offers a continuous measure of machine editing rather than a yes/no label \citep{thai2025editlensquantifyingextentai}. Across all of these approaches, the target of measurement remains the same -- how much of the final text's surface form was produced by a model. By asking how much of a document's content the human author contributed on top of what a model could already produce unprompted, VOLM answers a different question from what these methods are designed to answer.

Closest to our motivation is work by \citet{xie2026measuringhumancontributionaiassisted}, who also propose an information-theoretic measure of human contribution in AI-assisted content generation. They define human contribution as $\phi = I(x;y)/I(y)$, the ratio of mutual information between a human's input $x$ and the AI's output $y$ to the output's total self-information, where the self-information $I(y) = -\log p_\theta(y)$ is the output's probability under no conditioning at all, and the conditional self-information $I(y \mid x) = -\log p_\theta(y \mid x)$ is conditioned on the specific human input actually used to produce it.
Their method differs from ours in several key ways: first, their denominator, $I(y)$, measures surprisal relative to no context whatsoever, while VOLM measures surprisal relative to the task description $T$, which is the level of content freely obtainable by stating the task alone. Generation without context does not provide a useful proxy for zero human effort since even the least informative real request is still a task description, and comparing against no context at all sets a falsely permissive floor. Next, their primary measure requires the human input $x$, which an evaluator seeing only the finished document does not have. They propose an extension that doesn't require $x$ but this assumes any ``plausible" $x$ gives $y$ a per-token probability above a threshold $\tau$, yielding the lower bound $\hat\phi = 1 - \log(1/\tau)/\bar s(y)$, where $\bar s(y)$ is $y$'s average unconditional per-token surprisal. Their estimate is then a monotone function of the document's own perplexity, with $\tau$ hand-calibrated per model. VOLM instead conditions on $T$, compares against an explicit reference document $D_{\text{ref}} = M(T)$ rather than a scalar threshold, and never scores the document's surface text, only reconstructions of its extracted content, which insulates it from the stylistic variation that directly inflates a perplexity-based estimate. Beyond assigning a document-level score, VOLM also identifies the specific content that contributes to that score. \citet{rajendhran2026idealensdetectingaiideas} study the same question and train a classifier to detect AI ideas in long-form writing. Their classifier is trained on 1M FineWeb documents and uses labels from Pangram, a SOTA LLM-detector as the target. These are not gold labels and measure LLM-style/source provenance instead of idea provenance. In comparison, our method requires no additional model training or annotated data.

A separate line of work in HCI approaches contribution attribution from a perceptual or process-tracing perspective rather than an information-theoretic one. \citet{he2025contributionsdeservecreditperceptions} survey how knowledge workers believe credit should be assigned between human and AI collaborators across different types and amounts of contribution, finding that people consistently under-credit AI relative to an equivalent human partner.  \citet{kim2026ididntmakemicro} propose an operationalized framework that decomposes a task's goals into verifiable requirements and attributes each to the human or the model by tracing the full multi-turn dialogue. This requires access to the collaboration transcript, which VOLM does not require.

Finally, our framework is partially motivated by work that studies memorization by asking how much information is already encoded in a
language model's parameters. Most directly,
\citet{schwarzschild2024rethinkingllmmemorizationlens} propose the Adversarial Compression
Ratio, which measures the length of the shortest prompt that can elicit a target string from a model. A string is considered strongly memorized when a short prompt is enough to generate it. Our proposed framework shares this
compression-based intuition that content requiring
little additional information to reproduce is already
largely encoded in a model's weights. However, the two frameworks address different questions. Adversarial compression tests if a particular target string is stored in the model by optimizing a prompt that reproduces it exactly. VOLM instead evaluates a completed document without access to its original prompt, measures contribution relative to a task-specific baseline rather than an empty or optimized prompt, and does not require verbatim reproduction.

\section{Value Over Language Model (VOLM) Framework}
\label{sec:framework}

VOLM estimates how much of the content in a document $D$ originated with its human author, relative to a language model $M$ and a task description $T$. The central idea is to measure how much of the information content of the document must be removed before it is no more surprising to $M$ than what $M$ could produce from $T$ alone.

In order to operationalize this idea, we propose a pipeline consisting of 3 main parts: an Extractor $E$ that extracts the information content in the document in discrete units, a Selector $S$ that selects a subset of these discrete units to create an increasingly sparse reconstruction prompt, and a Reconstructor $R$ that reconstructs the document with the condensed information as context. Finally, we measure how surprised $M$ is when it sees this reconstructed document, i.e., we measure the surprisal (negative log-probability) of the text, given $T$. Figure~\ref{fig:volm-pipeline} illustrates the pipeline and we define each stage formally below.

\begin{figure}[t!]
\centering
\resizebox{\textwidth}{!}{%
\begin{tikzpicture}[
    box/.style={draw, rounded corners, align=center, font=\footnotesize, inner sep=2mm},
    io/.style={box, fill=gray!8, minimum height=10mm},
    mod/.style={box, fill=blue!5, minimum height=30mm},
    arr/.style={-{Latex[length=1.8mm]}, thick},
    fb/.style={-{Latex[length=1.8mm]}, thick, dash pattern=on 3pt off 2pt, gray!60!black},
    lbl/.style={font=\scriptsize, inner sep=1.5pt}
]
\node[io, text width=18mm] at (0,0)  (doc) {Document $D$\\Task $T$};
\node[io, text width=18mm] at (0,-2) (ref) {Reference\\$D_{\mathrm{ref}}=M(T)$};

\node[mod, text width=15mm] at (3.1,-1)  (E) {Extractor $E$\\[1mm]{\scriptsize $m$ samples}};
\node[mod, text width=20mm] at (6.5,-1)  (S) {Selector $S$\\[1.5mm]};
\node[mod, text width=17mm] at (10.1,-1) (R) {Reconstructor $R$\\[1mm]{\scriptsize $n$ per input}};
\node[mod, text width=11mm] at (13.2,-1) (M) {Scorer\\$M$};
\node[mod, fill=orange!8, text width=19mm] at (16.7,-1) (T) {};

\node[font=\footnotesize, align=center, anchor=north] at ([yshift=-1.5mm]T.north)
    {Distribution\\test};
\begin{scope}[shift={([yshift=10.5mm]T.south)}]
  \draw[gray, thin] (-0.95,0) -- (0.95,0);
  \draw[thick, teal]   plot[domain=-0.95:0.7, samples=40] (\x, {0.55*exp(-((\x+0.28)^2)/0.07)});
  \draw[thick, orange] plot[domain=-0.7:0.95, samples=40] (\x, {0.55*exp(-((\x-0.28)^2)/0.07)});
\end{scope}
\node[font=\scriptsize, align=center, anchor=south] at ([yshift=1.2mm]T.south) {$mn$ scores each\\[0.3mm]};

\draw[arr] (doc.east) -- (E.west |- doc.east);
\draw[arr] (ref.east) -- (E.west |- ref.east);
\foreach \a/\b in {E/S, S/R, R/M, M/T}{
  \draw[arr] (\a.east |- doc.east) -- (\b.west |- doc.east);
  \draw[arr] (\a.east |- ref.east) -- (\b.west |- ref.east);
}

\foreach \a/\b/\top/\bot in {E/S/g_K/g_{K,\mathrm{ref}}, S/R/g_i/g_{i,\mathrm{ref}},
                             R/M/\hat D/\hat D_{\mathrm{ref}}}{
  \coordinate (mid) at ($(\a.east)!0.5!(\b.west)$);
  \node[lbl, above] at (mid |- doc.east) {$\top$};
  \node[lbl, above] at (mid |- ref.east) {$\bot$};
}
\coordinate (mid) at ($(M.east)!0.5!(T.west)$);
\node[lbl, above, teal]   at (mid |- doc.east) {$\mathcal{S}_i$};
\node[lbl, above, orange] at (mid |- ref.east) {$\mathcal{S}_{i,\mathrm{ref}}$};

\draw[fb] (M.north) -- ++(0,0.65) coordinate (fbtop) -| (S.north);
\node[lbl, above, text=gray!60!black]
    at ($(S.north |- fbtop)!0.5!(fbtop)$) {scores guide the next deletion (greedy only)};

\coordinate (pv)  at ([yshift=-4mm]T.south);
\coordinate (swL) at ([shift={(-5mm,-9mm)}]T.south);
\coordinate (swR) at ([shift={(5mm,-9mm)}]T.south);
\draw[thick] (T.south) -- (pv);
\draw[thick] (pv) -- ($(pv)!0.8!(swL)$);
\draw[thick, gray, dash pattern=on 2pt off 1.5pt] (pv) -- ($(pv)!0.8!(swR)$);
\fill (pv) circle (1.3pt);
\fill (swL) circle (1.3pt);
\fill (swR) circle (1.3pt);
\node[lbl, left]  at (swL) {differ};
\node[lbl, right] at (swR) {no difference};

\draw[arr] (swL) -- ++(0,-0.3) coordinate (lp) -| (S.south);
\node[lbl, above] at ($(S.south |- lp)!0.5!(lp)$) {delete more content, $g_i \to g_{i-1}$};

\node[io, text width=17mm, anchor=north] at ([yshift=-0.6cm]swR) (out)
    {$\mathrm{VOLM}(D)$\\$=1-\frac{i}{K}$};
\draw[arr] (swR) -- (out.north);
\end{tikzpicture}}
\caption{\textbf{The VOLM framework.}
The extractor $E$ is run once ($m$ samples) on the document $D$ and task $T$, producing the full content representation
$g_K$; the reference $D_{\mathrm{ref}}=M(T)$ is processed identically (bottom lane).
At each step the selector $S$ deletes content to reach the next lower granularity $g_i$,
either greedily, using the scores of reconstructions (dashed feedback), or by a preset
schedule (no feedback). The reconstructor $R$ rebuilds a document $\hat D$ from $g_i$ and $M$ scores it,
giving score distributions $\mathcal{S}_i$ and $\mathcal{S}_{i,\mathrm{ref}}$, each pooling $mn$ scores ($m$ extractions $\times$ $n$ reconstructions).
A two-sample test compares them. While they differ, more content is deleted. When they
no longer differ, the procedure stops. With $\frac{i}{K}$ being the fraction of content
retained at that point, $\operatorname{VOLM}(D)=1-\frac{i}{K}$.}
\label{fig:volm-pipeline}
\end{figure}

\paragraph{Extractor $E$.} Given a document $D$ and task description $T$, an extractor $E$ extracts the information content of a document into a representation $g_K$. For instance, a simple 'fact-based' extractor extracts discrete facts or claims from $D$. Appendix \ref{sec:extractor} provides some other examples of extractors. Importantly, the choice of what to extract indicates what we value as the human contribution. Extraction can be a stochastic operation: in the above example, using an LLM to extract discrete claims from a document could lead to differences in the claims extracted across different runs. In our general framework, therefore, we assume the use of $m$ extraction samples, where $m$ is tunable.

\paragraph{Selector $S$.} Given the output of an extractor $g_K$, a selector $S$ constructs $g_i, i<K$, which are less informative representations of $D$. $g_K$ approximates the full content of $D$ and lower $i$ retain coarser, less specific information. For the aforementioned 'fact-based' extractor that extracts discrete facts/claims from $D$, a selector operates by choosing a subset of the discrete claims from $g_K$ instead of the whole set. Alternatively, consider an extractor that returns the document as is. A simple 'length-based' selector in this case is one that prompts a model to condense the document to a fraction of its original length, forcing the model to make the choice of retaining some information, and forgoing the rest.

\paragraph{Reconstructor $R$.} A reconstructor uses $M$ to generate a reconstructed document $\hat D_i = R(T, g_i, |D|)$ conditioned on the task description, the extracted content at level $i$ and the length of the original document $D$. Reconstructions are length-matched to $D$ (within 10\% of $|D|$, enforced as part of the generation prompt). Since reconstruction involves model inference, it can also be a stochastic process, and our general framework assumes the use of $n$ reconstructed samples, where $n$ is a tunable parameter.

\paragraph{Scoring.} The same evaluator model $M$ acts as the Scorer. Each reconstruction is scored by its average per-token log-probability under $M$, conditioned on the task description \emph{alone}:
\begin{equation}
\label{eq:score}
 s_i = \frac{1}{|\hat D_i|} \log P_M\left(\hat D_i \mid T\right) = \frac{1}{|\hat D_i|} \sum_{j=1}^{|\hat D_i|} \log P_M\left(\hat d_{i,j} \mid T, \hat d_{i,<j}\right),
\end{equation}
where $|\hat D_i|$ is the token length of $\hat D_i$ and $\hat d_{i,j}$ its $j$-th token. This asks how probable, on a per-token basis, $\hat D_i$ would be to a model that only ever saw the generic task description.

The pipeline only ever scores the reconstructions $\hat D_i$, which are always model-generated by construction. This ensures the comparison is less confounded by surface stylistic differences between human and machine prose. Length matching between $D$ and $\hat D_i$ is done so that differences in score cannot be attributed to differences in document length.

\subsection{The Replacement-Level Document}
\label{sec:reference}
Similar to VORP requiring a well-defined replacement-level player to serve as its baseline, VOLM requires a well-defined baseline reference document. We construct this baseline directly as
\begin{equation*}
    D_{\text{ref}} = M(T),
\end{equation*}
such that $D_{\text{ref}}$ is a document that $M$ produces from the task description alone, with no additional content supplied. We run both $D$ and $D_{\mathrm{ref}}$ through $E$, $S$, and $R$, and score the reconstructions identically, as described next.

\subsection{Statistical Test for Difference in Surprisal}
\label{sec:test}
As mentioned above, extraction and reconstruction can both be stochastic, giving two independent sources of variation at each information (granularity) level. For any input document $X$ (either $D$ or $D_{\mathrm{ref}}$), we therefore draw $m$ extraction samples $g_K^{(1)}, \dots, g_K^{(m)}$. The selector then produces the lower-granularity (lower-information) representations $g_i^{(j)} = S\left(g_K^{(j)}; i\right)$ for $i<K$, by filtering content from $g_K^{(j)}$. For each $g_i^{(j)}$ we draw $n$ reconstructions
\[
\hat X_i^{(j,l)} = R\left(T, g_i^{(j)}, |X|\right), \qquad
j = 1, \dots, m, \quad l = 1, \dots, n.
\]
Scoring each reconstruction as in Equation~\eqref{eq:score} yields, at each
level $i$, a distribution of $m \times n$ scores
\[
\mathcal{S}_i(X) = \left\{ s_i^{(j,l)} \right\}.
\]
The pipeline thus maps a document to a sequence of score distributions, one per granularity level. We write $\mathcal{S}_i = \mathcal{S}_i(D)$ and $\mathcal{S}_{i,\mathrm{ref}} = \mathcal{S}_i(D_{\mathrm{ref}})$.

VOLM measures the extent of content that needs to be removed from a document for the evaluator model $M$ to find it no more surprising than a replacement-level document. In practice, this is implemented through a statistical test.
At each granularity level $i$, we use a one-sided Mann-Whitney U test to compare $\mathcal{S}_i$ and $\mathcal{S}_{i,\mathrm{ref}}$. The alternative hypothesis is that reconstruction scores for the evaluated document tend to be lower than those for the reference, indicating greater surprisal under $M$. We reject the null hypothesis $H_0$ at $p<0.05$. A document that remains distinguishable even at coarse granularities requires substantial content removal and receives a high score while a document that becomes indistinguishable after little or no content removal receives a low score.

\subsection{Algorithm for Computing VOLM}
\label{sec:procedure}
Below, we lay out a general, step-by-step procedure for computing VOLM, agnostic of extractor choice. Concretely, suppose you are the evaluator, with access to $M$. You are given a document $D$ and its task description $T$.

\paragraph{Step 0: Generate the replacement-level document.}
Generate a reference document from the task description alone:
\[
D_{\mathrm{ref}}=M(T).
\]

\paragraph{Step 1: Extract.}
Construct the extracted version of the documents $g_K = E(D)$ and $g_{K,ref} = E(D_{ref})$.
Draw $m$ extraction samples from each document using $E$. Pair each sample from $D$ with one from $D_{\mathrm{ref}}$, starting with their full content representations.

\paragraph{Step 2: First reconstruction and scoring.}
For each representation, generate $n$ reconstructions using $R$ and score them as described in Section~\ref{sec:test}. Pooling scores across the $m$ extraction samples gives the document and reference score distributions $\mathcal{S}_i$ and $\mathcal{S}_{i,\mathrm{ref}}$. Test whether these distributions differ using the test described in Section~\ref{sec:test}. If the test does not reject, stop. Otherwise, reduce the content as detailed in the next step.

\paragraph{Step 3: Select content from the document.}
For each extraction pair, the selector $S$ reduces the document's current representation $g_i$ to the next lower granularity $g_{i-1}$, choosing from the candidate set $\mathcal{G}_{i-1}(g_i;E)$. How $S$ chooses depends on the selector. A Greedy Selector reconstructs and scores each candidate and selects the one with the highest mean score:
\[
g_{i-1}^{*}
=
\operatorname*{arg\,max}_{g\in\mathcal{G}_{i-1}(g_i;E)}
\bar{s}(g),
\]
where $\bar{s}(g)$ is the average score of reconstructions generated from $g$. A Preset Selector instead follows a fixed deletion order (such as in reverse order of the fact's appearance in the document) that does not depend on any score. Greedy selection can be computationally expensive, since every candidate must be reconstructed and scored at every step, so preset schedules might be better suited to certain applications.

\paragraph{Step 4: Select content from the reference.}
Similarly, $S$ reduces the reference representation to the next lower granularity. Under greedy selection, it picks the candidate whose mean reconstruction score is closest to that of the updated document representation:
\[
g_{i-1,\mathrm{ref}}^{*}
=
\operatorname*{arg\,min}_{
g\in\mathcal{G}_{i-1}(g_{i,\mathrm{ref}};E)}
\left|
\bar{s}(g)-\bar{s}(g_{i-1}^{*})
\right|.
\]
Under a preset selector, the reference follows the same deletion order as the document. Repeat Steps 2--4, reconstructing, scoring and comparing at each new level, until the stopping condition is met.

\paragraph{Step 5: Report.}
Let $i^*$ be the level at which the test first fails to reject, so that the document retains a fraction $i^*/K$ of its content. Define
\[
\operatorname{VOLM}(D)=1-\frac{i^*}{K}.
\]

\section{Experiments}
\label{sec:experiments}
We conduct experiments and ablation studies on datasets spanning journalism, academic peer review, and argumentative essay writing, to show our proposed algorithm's efficacy in identifying original contribution across domains with meaningfully different notions of a task description $T$. 

\subsection{Data} 
\label{sec:data}
\paragraph{\textbf{All the News:}} We randomly sample 100 articles from the \textit{All the News} \citep{thompson2020allthenews_hf} corpus. $T$ is the article's published headline.

\paragraph{\textbf{ICLR Reviews:}} We randomly sample 200 papers from ICLR 2023, selecting one review at random per paper. $T$ consists of the full paper, the venue's reviewer instructions, and its rating scale.

\paragraph{\textbf{Persuasive Argumentation:}} We randomly sample 200 documents from the 8 independent-writing prompts in the PERSUADE 2.0 corpus \citep{CROSSLEY2024100865} (25 essays per prompt). We exclude the corpus's 7 source-based prompts, which would require source material as part of $T$. $T$ is then the essay prompt itself.

In each domain, we begin with a set of human-written documents and construct a matched LLM-generated counterpart for each, by prompting $M$ with the same $T$ as the human document. For every document, its LLM-generated counterpart is constrained to be within 10\% of the human document's length, enforced as part of the generation prompt, to remove length as a confounder. More prompting details are provided in Appendix \ref{sec:app}. We confine the results in the main text to the All the News dataset, and defer the same results for other datasets to Appendix \ref{sec:iclrresults} and \ref{sec:persuaderesults}.

\subsection{Method Instantiation}
\label{sec:instantiation}

The framework in Section~\ref{sec:framework} is agnostic to the specific choice of extractor $E$, selector $S$, reconstructor $R$, and model $M$. For the experiments that follow, we fix a concrete instantiation, described here for the All the News dataset.

\paragraph{Model}
We use Llama 3.1 8B Instruct \citep{grattafiori2024llama3herdmodels} as $M$ throughout, serving both as the extractor's underlying model and as the reconstructor.

\paragraph{Extractor}
For the experiments in the main text, $E$ extracts an ordered list of discrete facts from the document. We define a fact as a record with a fixed schema. This consists of a subject, an action, an object, and any associated quantity, date, or quotation. We constrain facts in this way to limit how much of the original wording can pass through extraction and reconstruction because stylistic carryover would otherwise appear as an information difference. For the same reason, we restrict the action to a base-form verb canonicalized with WordNet: when several senses are available, we select one using the deterministic Lesk algorithm \citep{10.1145/318723.318728}, with the fact's subject and object as context, and render the fact with that sense's representative verb. In practice, we prompt $M$ to decompose the document into the schema and then apply the canonicalization. We refer to this as the Equalized Schema-Constrained Fact Extractor with WordNet canonicalization and provide prompting and schema details in Section~\ref{sec:bestextractor}. ``Equalized'' means that the number of facts extracted from the reference within each (document,reference) pair is used as the target count for the document.

\paragraph{Selector}
We test two selectors. Both operate on the ordered list of $K$ facts produced by the extractor, so $g_K$ is the full list. In general, $g_i$ is itself an ordered list of $i$ facts, and every selection step takes $g_i$ to $g_{i-1}$ by deleting exactly one fact. $D$ and $D_{\mathrm{ref}}$ both start from $K$ (variable) facts and are reduced one fact at a time.
\begin{itemize}
    \item \textbf{Preset (ordered).} This selector deletes the fact that appears last in the list, so that $g_i$ consists of the first $i$ facts, i.e., facts are removed in the reverse order of their appearance in the document.
    \item \textbf{Greedy.} This follows Steps 3 and 4 of Section~\ref{sec:procedure}. Greedy selection is an informative selector because the order of removal reflects each fact's contribution to surprise. However, it can be computationally expensive, requiring $O(K^2)$ candidate evaluations per extraction sample in the worst case, compared to $O(K)$ for the preset selector.
\end{itemize}

\paragraph{Reconstructor}
$R$ prompts $M$ to write a document satisfying $T$, given the facts $g_i$ included at that level as additional context, and constrained to fall within $10\%$ of the original document's word count, matching the length-matching described in Section~\ref{sec:data}.

\subsubsection{Procedure}
\label{sec:procedure2}
We instantiate the procedure of Section~\ref{sec:procedure} for All the News as follows: $T$ is the article's published title, and $E, S$ and $R$ are as defined above. The selector still deletes content one fact at a time. The distribution test is run only at eleven fixed checkpoints, defined by the fraction of the extracted facts retained, $\gamma \in \{1.0, 0.9, \dots, 0.0\}$. At checkpoint $\gamma$, extraction sample $j$, which has $K_j$ extracted facts, retains $\lceil \gamma K_j \rceil$ of them. Once every sample has been reduced to this count, we generate $n$ fresh reconstructions from each, score them, and pool the scores across the $m$ extraction samples to run the test in Section~\ref{sec:test}. We define checkpoints for evaluation, rather than evaluating after every fact deletion, because different extraction samples can extract a different number of facts from the document. Fixed checkpoints allow us to compare and report VOLM on a common scale across them. At each checkpoint, we can then use data from every extraction path for the distribution test. Starting from $\gamma = 1.0$, we move to the next checkpoint while the test rejects, and stop at the first checkpoint $\gamma^*$ at which it does not, giving $\operatorname{VOLM}(D) = 1 - \gamma^*$. If the test rejects at every checkpoint, including $\gamma = 0.0$, then $\operatorname{VOLM}(D) = 1.0$; if it does not reject at $\gamma = 1.0$, then $\operatorname{VOLM}(D) = 0.0$. As a result, VOLM scores take values in $\{0.0, 0.1, \dots, 1.0\}$, which also means that multiple documents share the same score.

\subsection{How does a single document behave under the VOLM pipeline?}
\label{sec:exp-single-doc}
To demonstrate how an individual document behaves under the VOLM pipeline, we present the score-versus-granularity curve for a representative article and its replacement-level counterpart, using the Greedy Selector. Using a Greedy Selector gives us, in addition to a VOLM score, the list of facts that $M$ found most surprising (the facts removed before the procedure stops). Figure~\ref{fig:two-sides}a plots the average per-token log-probability of reconstructions against the fraction of extracted facts retained (using $m=n=8$). The reference curve is mostly flat, since all of its facts were extracted from a document that $M$ itself produced from $T$, so supplying more or fewer of them hardly changes how surprising the reconstruction is. The human article's curve starts well below the reference and rises as the most surprising facts are removed. The procedure yields a VOLM of 0.7. Appendix~\ref{sec:single-doc-facts} contains the specific ``high-VOLM'' facts that the Greedy Selector identifies as the ones the model found most surprising. This is a result on only a single article, however, and in the next section, we extend it to the whole corpus.

\subsection{How do VOLM scores differ between human and LLM-written articles?}
We compare VOLM scores across two sets of articles: human-written articles, and LLM-generated articles that are also produced using $M$ and prompt $T$, but are different from the reference articles. Figure~\ref{fig:two-part}a shows the distribution of VOLM scores for the two sets using the Greedy Selector. We find that the LLM-generated articles have a median VOLM of 0.0 (mean=0.071), while the human-authored articles have a median VOLM of 0.7 (mean=0.673).

We also present the same figures for a Preset Selector in Figure~\ref{fig:volmdistappleft} in Appendix~\ref{sec:volmdistapp}. VOLM scores measured using the two selectors are significantly and positively rank-correlated (Kendall's $\tau=0.265$, $p=0.00141$ for human-authored articles and $\tau=0.272$, $p=0.00264$ for LLM-authored articles). We therefore present Preset Selection as a promising, easy-to-run alternative to Greedy Selection, useful where quick evaluation matters more than precision. We conduct the remaining experiments in the main text using the Preset Selector, with the corresponding Greedy results in Appendix \ref{sec:populationgreedyapp}.

\subsection{How do human-written articles behave under the VOLM pipeline?}
\label{sec:exp-population}
We use population-level validation to show how human-ideated documents behave when subjected to our pipeline, and how their score curves diverge from the reference curve. We use a corpus-level comparison for this initial illustration, rather than a single-document comparison, so that variation between individual documents averages out instead of tying our results to the particulars of any one document. This also mirrors \citet{liang2024mappingincreasingusellms}, who studied AI detection at a corpus level, identifying the amount of LLM use in paper abstracts. Because we rely on the variance across documents, we set $m=n=1$, using a single extraction and a single reconstruction per document.

\begin{figure}[t!]
\centering
\begin{subfigure}[t]{0.32\textwidth}
    \centering
    \includegraphics[width=\linewidth]{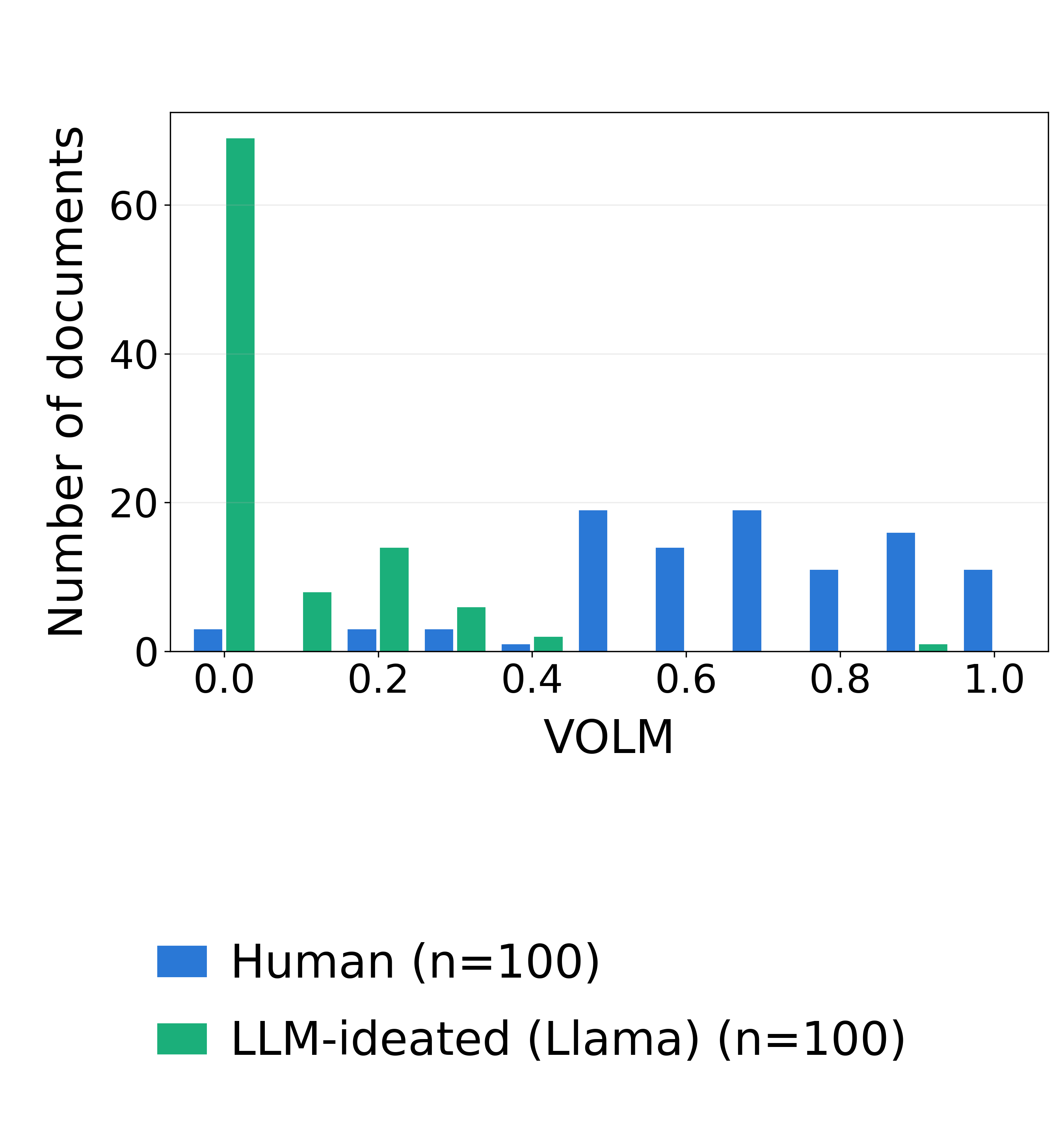}
    \label{fig:greedyvolmdist}
\end{subfigure}\hfill
\begin{subfigure}[t]{0.66\textwidth}
    \centering
    \includegraphics[width=\linewidth]{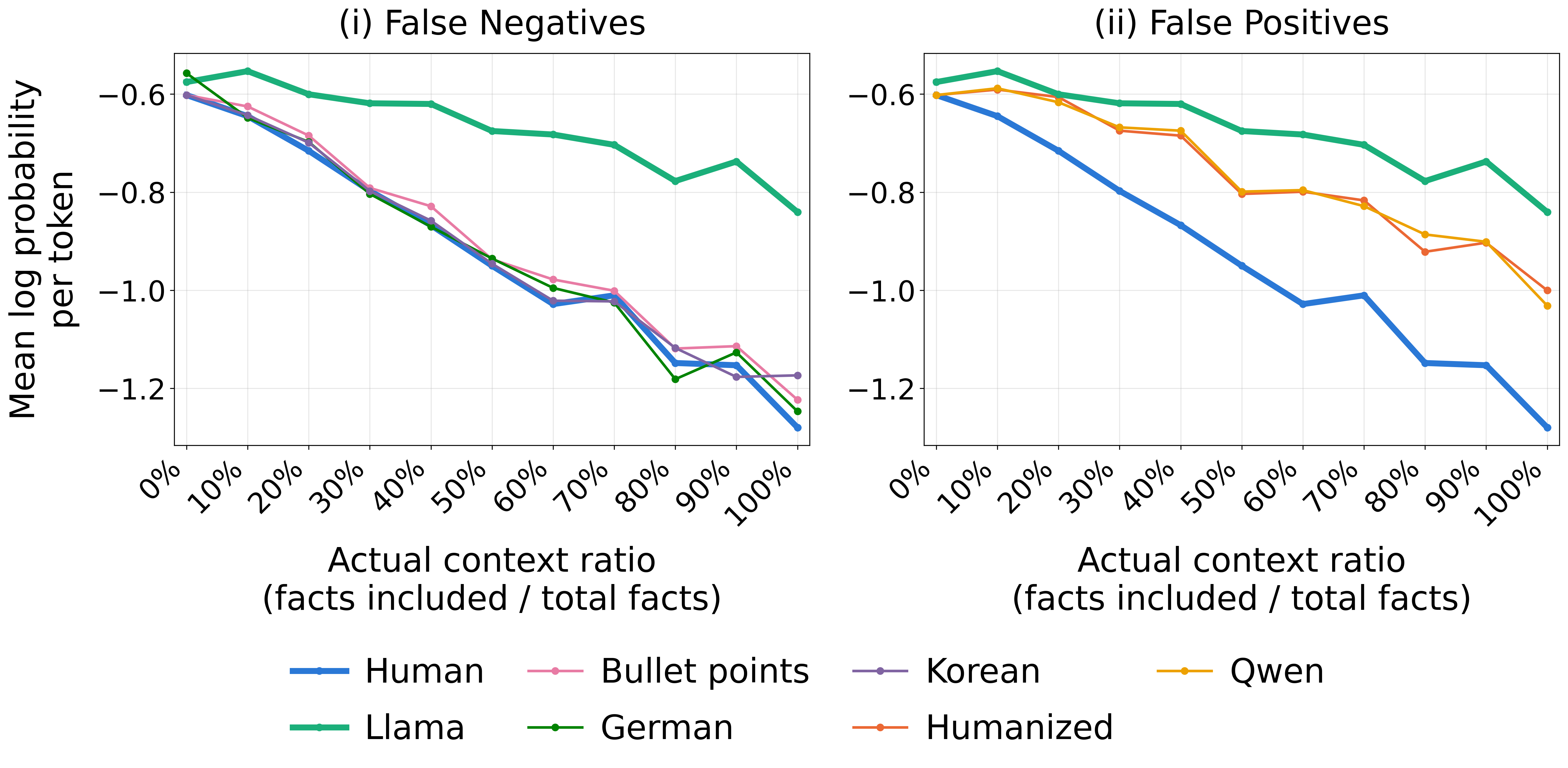}
    \label{fig:populationnews}
\end{subfigure}
\caption{ (a) Left -- Distribution of VOLM scores for human-written articles from the All the News dataset and LLM-generated articles. (b) Right -- Averaged log-probability curves on the All the News dataset. (i) Curves for the reference LLM (Llama)-ideated corpus, human-ideated corpus, and LLM-transformed variants of the human-ideated corpus. (ii) Curves for the reference LLM-ideated corpus, human-ideated corpus, Qwen-ideated corpus and humanized corpus. }
\label{fig:two-part}
\end{figure}

For every document $d$, we run the pipeline at the eleven retention checkpoints $\gamma$ of Section~\ref{sec:procedure2}. As content is removed, the human scores should rise toward the reference scores, since what remains increasingly resembles what $M$ could produce from $T$ alone. The reference scores should change less, because all the information retained for the reference was itself extracted from an article generated by $M$ from $T$ alone. Figure \ref{fig:two-part}b(i) shows this (human curve is bolded blue and reference Llama curve in bolded green).

\subsection{Is VOLM robust to LLM-style writing? (False Negatives)}
\label{sec:exp-hierarchy}
A stronger test of VOLM is whether it tracks a document's information content specifically, rather than superficial properties of its authorship. As stated earlier, the goal of our method is to detect original contribution in writing, on top of what a language model could have easily produced, and a document being written in 'LLM-style' should not affect this measurement. 

We construct, for each document of each dataset, two variants:
\begin{itemize}
    \item \textbf{Bullet-reconstructed:} the article is condensed to bullet points, then reconstructed into prose using a language model. We use $M$ itself, as a worst case scenario: the text will now have been subjected to two LLM transformations using the same model that is being used to evaluate it and is now ostensibly written in the style of $M$.
    \item \textbf{Round-trip translated:} the article is translated to another language and then back to English, both using $M$. We do this for two languages: German and Korean, so as to have a sample in the same language family as English and another in an entirely different language family that could force the LLM to not retain the structure and phrasing of the original text while translating.
\end{itemize}
Both variants are, by surface appearance, heavily LLM-touched and a detector measuring surface LLM involvement could plausibly flag either as machine-written. Neither, however, introduces or removes any information relative to the original article. Passing each variant through the procedure of Section~\ref{sec:instantiation}, we expect both to produce score curves close to the original human article's curve, and clearly separated from $D_{\text{ref}}$'s, since VOLM importantly never scores a document's own surface text and only ever scores reconstructions built from its extracted content.

We compute similar group-averaged score curves for these 3 variants. Figure \ref{fig:two-part}b(i) shows these curves for the All the News dataset. Each of the three curves closely tracks the human curve and takes a while to meet the Llama curve, indicating the presence of original content beyond what the language model could easily write. Subjecting original text to LLM transformations did not alter its evaluation by our method.

\subsection{Can VOLM be fooled by other LLMs or humanizers? (False Positives)}
\label{sec:robust}
We check VOLM's robustness to two adversarial settings below, namely, documents written by a language model different from the evaluation model $M$, and LLM-written documents that have been passed through a humanizer. an ideal measure of VOLM is one that is resistant to stylistic confounding and solely measures information gaps, i.e., whether the presented document contains information that a language model would not have generated with a short task description prompt. Ideally, both settings have a low VOLM score, since they did not involve substantial human contribution on top of what a language model could've easily written with a generic prompt. 

\paragraph{Text written by other language models}
We use the same task description prompt $T$ to generate a document using a different language model from $M$. In our test instantiation, we use Qwen 3-8B \citep{qwen} as the generator and Llama 3.1 8B as the evaluator model $M$.

Figure \ref{fig:two-part}b(ii) shows the averaged log probability curve following the procedure detailed in Section \ref{sec:procedure2} for Qwen generated documents in yellow.

\paragraph{Humanization}
Humanizers are designed to evade LLM detectors by adversarially paraphrasing text in a manner that bypasses trained detectors. Though our method is not an LLM detector and measures a different objective (the value provided on top of a language model), robustness to humanization is still important -- a piece of text should not acquire a higher VOLM score, simply because it was paraphrased by a humanizer.

We use DIPPER \citep{krishna2023paraphrasingevadesdetectorsaigenerated} as the humanizer and apply it to our reference corpus of Llama generated documents.

Figure \ref{fig:two-part}b(ii) shows the averaged log probability curve following the procedure detailed in Section \ref{sec:procedure2} for humanized documents in orange.

The figure shows a clear hierarchy -- the reference curve, made up of documents generated by $M$ itself from $T$ alone, increases only slowly as we reduce the number of facts provided as context. This is because $D_{\text{ref}}$ contains no information $M$ did not already supply, and the addition/deletion of facts from $D_{\text{ref}}$ does not lead to substantial surprisal. The other-model and humanized curves form a second tier. They are steeper than the reference curve but remain well above the bottom tier. One was generated by a different model and the other adversarially paraphrased, so both differ from the reference in surface origin, but neither introduces any genuinely new information, and VOLM correctly places both above documents with real original content. The bottom tier contains the human curve together with its two content-preserving transformations from Section~\ref{sec:exp-hierarchy}, the bullet-reconstructed and round-trip translated variants. All three increase sharply as facts are deleted and closely track one another, despite having been substantially rewritten by an LLM at some stage, since none of these transformations removes the information the human author originally supplied.

\begin{figure}[t!]
\centering
\begin{subfigure}[t]{0.33\textwidth}
    \centering
    \includegraphics[width=\linewidth]{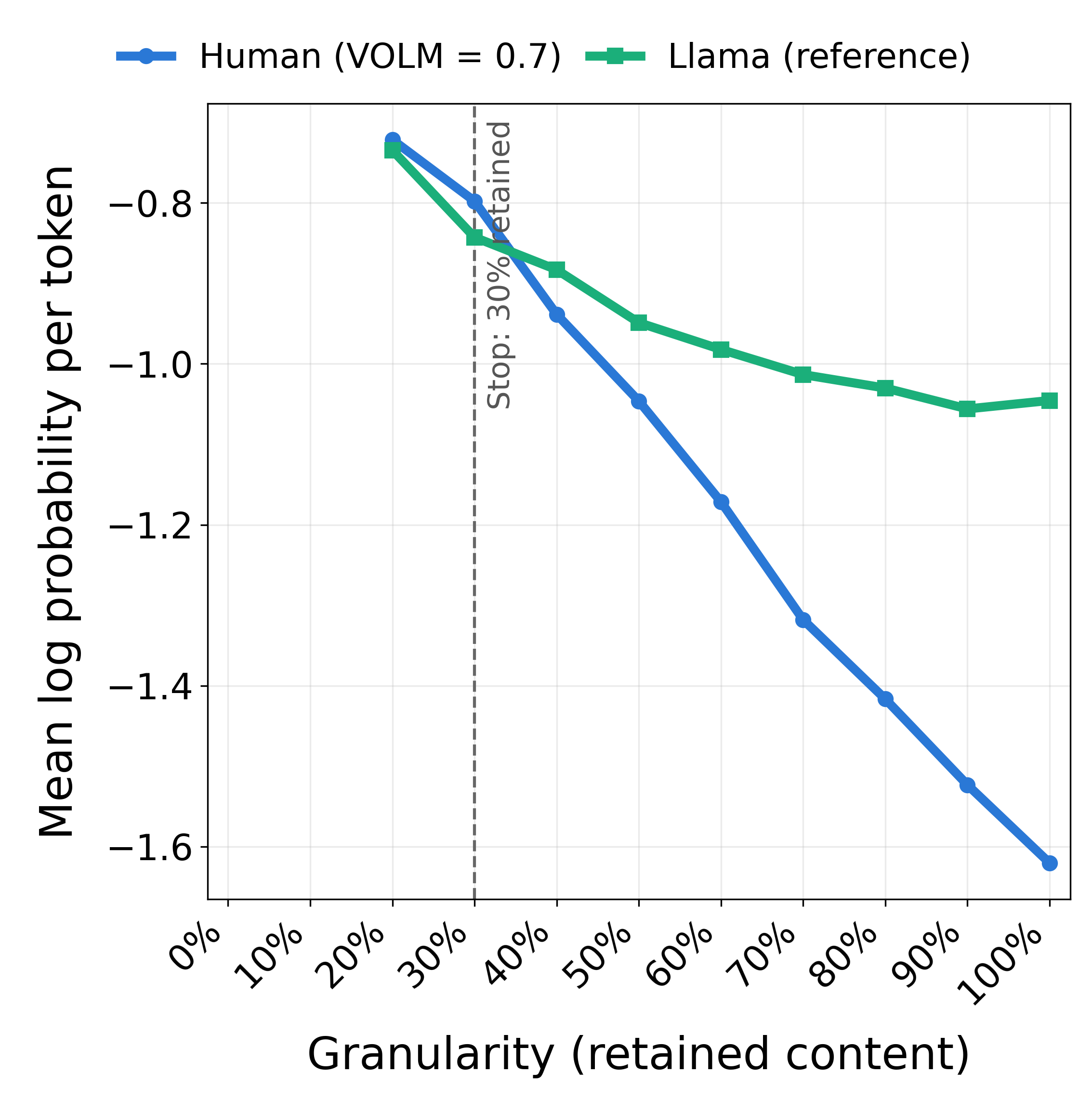}
    \label{fig:single}
\end{subfigure}\hfill
\begin{subfigure}[t]{0.66\textwidth}
    \centering
    \includegraphics[width=\linewidth]{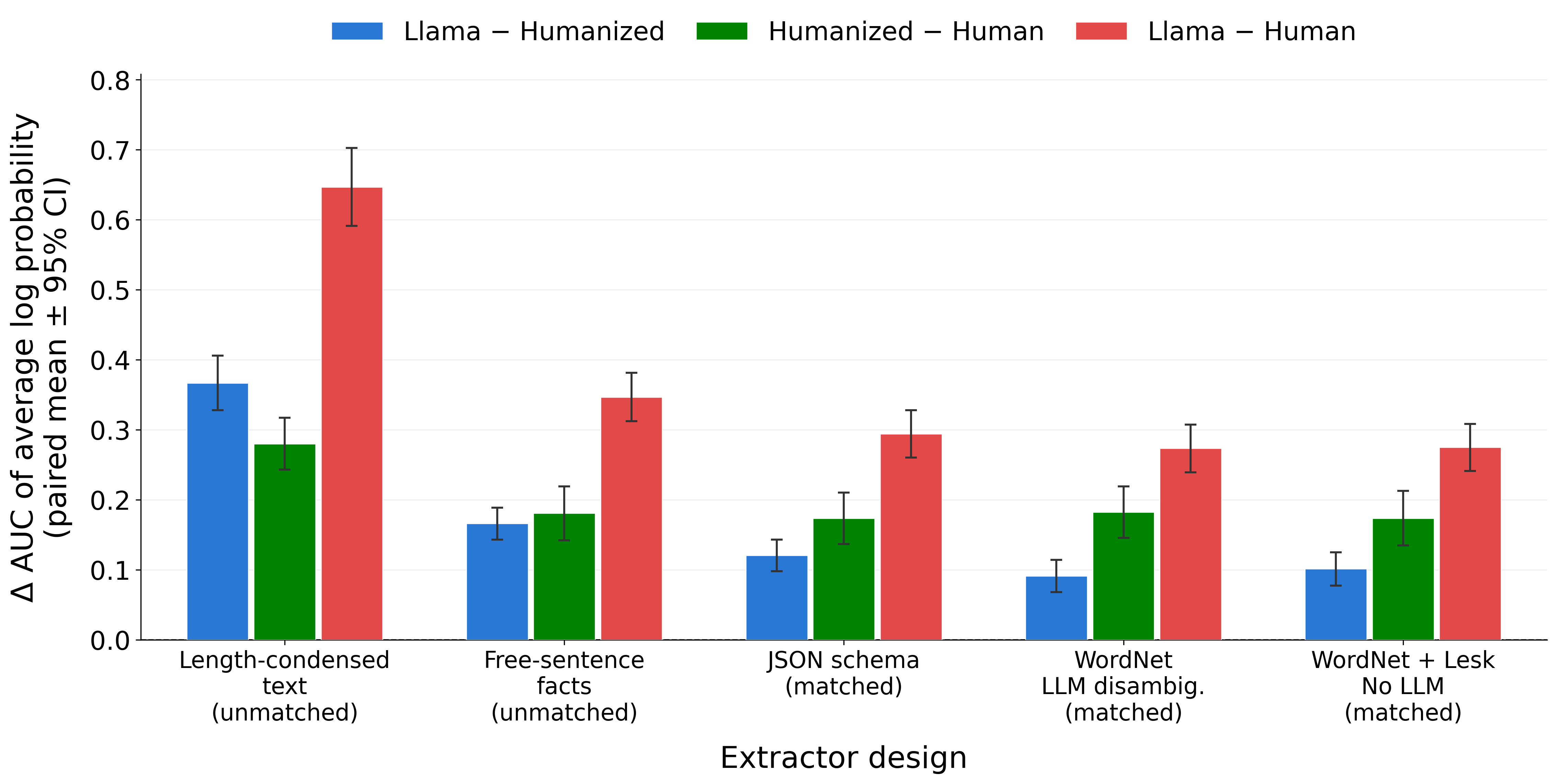}
    \label{fig:extractorcomparison}
\end{subfigure}
\caption{(a) Left -- Averaged log-probability curves for a sample document from the All the News dataset. (b) Right -- Paired AUC gap mean of avg log-probability per token, integrated over the compression sweep, paired per document, (n=100) between each source pair, tracked across successive extractor designs detailed in Appendix \ref{sec:extractor}. A sharper extractor makes humanized text look more like unconditioned LLM output.}
\label{fig:two-sides}
\end{figure}

\subsection{How does the choice of extractor matter?}
\label{sec:exp-extractor}
The above experiment shows that text from a different model and humanized text receive a non-zero VOLM, falling between the reference and human curves even though neither introduces substantial new information (a model substantially different from the evaluator $M$ could introduce 'new' information). A likely explanation could be the persistence of residual stylistic information. Text from other models or a humanizer carries different token-level statistics from $M$'s own output, which lowers its log-probability under $M$, even if it has similarly easy to generated ideas. To explore this, we consider methods to mitigate style retention, and experiment on better extractor design to retain maximum information while ignoring style artifacts. We begin with the identity extractor and a length-based selector that prompts $M$ to condense the document to a fraction of its original length, but find that word choices from the original text persist through condensation and reconstruction (Appendix \ref{sec:wordsretained}). We therefore test whether more constrained extractors reduce this effect.

To summarize a document's whole logprobability vs granularity curve as a single scalar, we integrate its average log-probability over the retention sweep. For two sources $A$ and $B$, we define the gap between them as the paired difference in AUC, $\mathrm{AUC}_A(d) - \mathrm{AUC}_B(d)$, computed per document $d$ and averaged over the corpus ($n=100$) with a 95\% confidence interval. Pairing on the same document cancels document-level difficulty (some titles are inherently easier or harder to predict than others) and isolates the effect of the source itself.

We compare extractors with increasingly rigid extraction parameters. Appendix ~\ref{sec:extractor} specifies these extractors in more detail. Figure ~\ref{fig:two-sides}b shows that as we constrain the extractor, the gap between the Llama and humanized corpora narrows. Moving from the condensation-based extractor to the fact extractor, the schema-constrained equalized fact extractor, and finally the WordNet-canonicalized equalized schema extractor, the Llama-Humanized gap fell from 0.367 to 0.166 to 0.121 to 0.101, a total reduction of 72.4\%. Substituting an LLM-based word-sense disambiguation step for Lesk \citep{10.1145/318723.318728} gave roughly the same result. By contrast, the Humanized-Human gap remained essentially unchanged (0.181, 0.174, and 0.174 for the fact, equalized schema, and equalized schema with WordNet extractors respectively, a net change under 4\%). Only comparisons with the reference LLM were affected by the extractor, since log-probabilities are measured with respect to the reference LLM.

\section{Discussion}
The results in Section \ref{sec:experiments} paint an interesting picture. The idea behind VOLM is principled: in order to figure out the amount of information content in a document that is 'genuinely new' and could not have easily been produced by a language model without deliberate prompting, we measure how surprised a language model would be to see the document if it only had a generic prompt to work with. However, this is very prone to stylistic confounding (LLMs would be less surprised to see text that is ostensibly 'LLM-style'), so this affects a straightforward log-probability based metric. In order to avoid this, we 1) ensure every document scored by the evaluator model $M$ is a reconstruction via $M$ itself, and not original text, and 2) iterate on better extractors that strip away style. We expand on the second point below, building on the extractor experiment of Section \ref{sec:exp-extractor}.

We show that increasingly constrained extractors narrow the gap between Llama-generated and humanized text, while leaving the gap between humanized and human text mostly unchanged. This provides evidence that the reason for non-zero VOLM scores for some humanized text is because of persistent stylistic confounding and that more constrained extractors can partially resolve this while retaining important information. We therefore encourage more involved work on extractor design within our proposed framework, so that VOLM measures information gaps and not stylistic differences.

The instantiation of our framework used for the experiments in this paper also presently attributes a non-zero VOLM to text produced by models different from the evaluator model $M$, though the magnitude of this VOLM is generally lower than it is for genuinely original documents. Intelligent thresholding or defining what VOLM level is significant could resolve this issue. Another way to get past this is to use multiple evaluator models $M$ and measure VOLM for all of them. We note that this is an artifact of the method itself -- small differences in training data and styles (if the extractor isn't optimized) could lead to one model seeing the text generated by another model as original to some degree.

A key motivating idea for VOLM came from a thought experiment: An experienced computer science researcher asked to condense a computer science document can easily choose what to keep, because she has the domain knowledge to express its ideas concisely. When asked to condense a biology document, she would struggle to choose what to keep/discard. Along similar lines, we posited, ``An LLM will find it harder to condense information that isn't already implicit in its weights." When we design a selector $S$ to approximate the original document $D$ at different granularity levels, we are performing the same information compression. If the document that an LLM generates when it is provided a large fraction of the information in a given document is not substantially different than what it would've generated when not provided with any of that information, then the document's author did not provide substantial input that influenced the creation of the final document, implying a low VOLM.

VOLM solves a different problem from traditional LLM detection, and its use cases are accordingly different, and sometimes complementary. An interesting observation from our results is that VOLM scores observed on the PERSUADE 2.0 Essay dataset are lower than those for the All the News dataset, and we interpret this as the lack of lesser-known facts used by middle/high school students in essay writing as compared to News articles. We use this observation to assert the need to be careful in choosing areas where VOLM is/is not appropriate as a metric. Peer review in particular was a strong motivating area in the ideation of this framework. With VOLM as a guiding metric, reviewers are encouraged to provide diverse criticisms, different from what an LLM could easily produce by itself, thus encouraging them to contribute their own insight about the paper instead of restating what a model would say anyway. If a fully LLM-generated review is already provided to authors, human reviewers can then focus on identifying different issues, and we can evaluate their reviews as such. In the worst case, if every reviewer submitted the one-shot output of an LLM, they would come up with the same criticisms and provide less useful signal about the paper's quality.

\section{Conclusion}

As people continue to adopt LLMs into their writing process and LLMs become increasingly capable of complex reasoning, authorship attribution is becoming both more ambiguous and more important. The question is no longer just whether a person used an LLM, or even how much of a document's final text was generated by one. An author may rely heavily on an LLM to render a document's prose while still supplying the ideas, arguments, and structure that give it its substantive value. Conversely, a document could be the product of little or no original human contribution even if its surface form has been modified to appear more human-written. In this paper, we introduce a training-free, annotation-free framework for making this distinction. Instead of measuring the fraction of text attributable to an LLM, VOLM measures a document's 'value' relative to what a language model could already produce from the task description alone. VOLM aims to quantify how much information in a document goes beyond what was already readily available to the model, serving as a proxy measure of idea provenance. We note and emphasize that ``value'' in VOLM does not refer to the intrinsic merit, usefulness, or social significance of a document's ideas. 'Value' and 'contribution' are both loaded terms and we do not assert that our measure captures a subjective notion of either of these terms. Instead, it specifically captures how difficult those ideas would be for a language model to produce from the task description alone. Content that is less readily produced by the model has higher VOLM, regardless of its broader worth. A document composed of ideas and claims that are not generated by a language model with high probability is identified as high VOLM and documents containing claims that are easily reproduced by a language model are low VOLM.

Our experiments demonstrate the feasibility of this approach across journalism, peer review, and argumentative essays. VOLM separates human-ideated documents from documents generated from generic task descriptions and identifies document claims that would have been the hardest for the LLM to come up with, while remaining substantially invariant to content-preserving transformations, including LLM reconstruction and round-trip translation. At the same time, our analysis discovers that residual stylistic information can often persist even after multiple rounds of LLM-assisted rephrasing operations, causing text produced by different models or adversarial humanizers to receive non-zero VOLM scores despite containing little additional original information. We show that increasingly constrained extractors can substantially reduce this effect, pointing to extractor design as an important direction for future work.

We view the present work as a first step toward measuring contribution at a granular level rather than merely detecting AI use. More principled extractors and reconstructors and multiple evaluator models are all promising directions for future work. As the boundary between human and machine writing continues to blur, methods that distinguish who produced the words from who contributed the ideas may become increasingly important for assigning credit, evaluating work, and understanding what relevant human contribution looks like in the age of language models.

\section*{Generative AI usage statement}
Generative AI tools have been used to assist in formatting \LaTeX (creating tables and improving alignment).

\section*{Ethics and Privacy Statement}
We use this section to reiterate two important points to keep in mind when applying VOLM. Instead of asking how much of a text a model produced, VOLM offers an alternative regime of evaluating LLM-assisted writing, by asking how easy it would have been for a model to generate the ideas in the text from the task description alone. This viewpoint is relevant for writers, including differently abled and non-native speakers, who use LLMs to render their own ideas into prose, as LLM involvement in the surface text is not treated as a signal in itself. We do not propose VOLM as a tool for gatekeeping or assessment, and the following two points explain why:
\begin{itemize}
    \item VOLM is a statistical estimate relative to a model $M$ and a task description $T$. A high or low score as should not be used as proof of authorship or as a tool for penalization. Humans can write a document with ideas that are low VOLM, i.e., unoriginal with respect to an LLM.
    \item VOLM depends on what $M$ finds surprising, so it inherits $M$’s biases. Writing from underrepresented topics might score differently for reasons unrelated to original contribution.  
\end{itemize}

\bibliographystyle{ACM-Reference-Format}
\bibliography{bibliography}

\appendix
\section{Appendix}
\label{sec:app}

\subsection{Extractor Design}
\label{sec:extractor}
\subsubsection{Identity Extractor}
\label{sec:condensation}
An identity extractor simply returns the document itself as its  information-extracted version $g_K$. It can be paired with a condensation-style selector that constructs a document $g_i$ at granularity level $i$ by prompting the evaluator model $M$ to condense the document to a fraction $i/K$ of its original length. Condensing the document to a shorter length forces $M$ to retain some information, and let go of the rest, thereby functioning as a proxy for reducing information content.

\subsubsection{Fact Extractor}
A fact extractor first prompts the evaluator model $M$ to extract facts from the provided document with no limit mentioned. 

It can then be paired with a selector that constructs a document $g_i$ at granularity level $i$ by only taking some $i/K$ fraction of those facts. For instance, if the extractor extracted 10 facts from the document, $g_1$ would only consist of one fact. 

\subsubsection{Equalized Fact Extractor}
An equalized fact extractor is the same as a generic fact extractor in all respects, except it sets a target on the number of facts to be extracted. By prompting, we ensure that the number of facts extracted from any document that is being evaluated is the same as the number of facts that were extracted from the reference LLM document.

\subsubsection{Equalized Schema-constrained Fact Extractor}

An equalized schema-constrained fact extractor is an equalized fact extractor with the additional constraint that each extracted fact must conform to a fixed schema. Specifically, each fact is represented using the following fields:

\begin{table}[h]
\centering
\begin{tabularx}{\linewidth}{lX}
\hline
\textbf{Field} & \textbf{Description} \\
\hline
\textit{subject} & Entity central to the fact, given exactly as named in the source. \\
\textit{action} & Relation or event described by the fact. \\
\textit{object} & Remainder of the claim, preserving all named entities. \\
\textit{quantity} & Number, percentage, or amount associated with the fact, if any. \\
\textit{date} & Date, year, or time reference associated with the fact, if any. \\
\textit{quote} & Direct quotation associated with the fact, if any. \\
\hline
\end{tabularx}
\caption{Schema used by the schema-constrained fact extractor.}
\label{tab:fact-schema}
\end{table}

For example, given the following text:

\begin{quote}
Ford Motor Company announced plans in 2020 to invest 1.5 billion dollars in its plant in Chongqing to produce the Ford Mondeo.
\end{quote}

the extractor produces the following structured fact:

\begin{verbatim}
{
"subject": "Ford Motor Company",
"action": "invest",
"object": "in its plant in Chongqing to produce the Ford Mondeo",
"quantity": "$1.5 billion",
"date": "2020",
"quote": null
}
\end{verbatim}

The schema constrains both the information represented by each fact and the manner in which that information is subsequently serialized. Now, the extracted records are deterministically rendered using a fixed template rather than being rewritten as free-form sentences by the language model.

\subsubsection{Equalized Schema-Constrained Fact Extractor with WordNet Canonicalization}
\label{sec:bestextractor}
We first test whether constraining the action field to a fixed
vocabulary improves the consistency of the extracted facts. We use a vocabulary of 61 canonical action verbs, with an other:<lemma> escape hatch for relations that do not fit the predefined vocabulary. This intermediate extractor reduces stylistic variation relative to the schema-constrained extractor alone.

We then take this idea further by removing the fixed-vocabulary restriction and instead canonicalizing actions using WordNet. The extractor first identifies a base-form verb that best describes each relation or event. For each extracted verb, we retrieve its possible WordNet verb senses and use the deterministic Lesk algorithm \citep{10.1145/318723.318728} to select the sense that best matches the fact's context. The canonical action is then obtained from the selected WordNet synset and used when rendering the fact. We also try using a language model (the same evaluator model $M$) as a word-sense disambiguator to choose the correct base-form verb and achieve similar results to when we use Lesk's algorithm.

The fixed-vocabulary experiment therefore serves as an intermediate test of whether restricting action vocabulary is beneficial, while the WordNet version generalizes this approach beyond a manually specified set of actions. All experiments reported in the main text use the latter WordNet-canonicalized extractor.

\subsection{Results on ICLR Reviews}
\label{sec:iclrresults}
Figure \ref{fig:iclrpersuade}a shows the logprobability curves for the ICLR 2023 review dataset. The extractor $E$ here is the simple identity extractor described above in Section \ref{sec:condensation}, paired with a condensation-style selector. The reference LLM (Llama) curve stands out against every other curve.

\subsection{Results on Persuade Essays}
\label{sec:persuaderesults}
\begin{figure}[t]
    \centering
    \includegraphics[width=\textwidth]{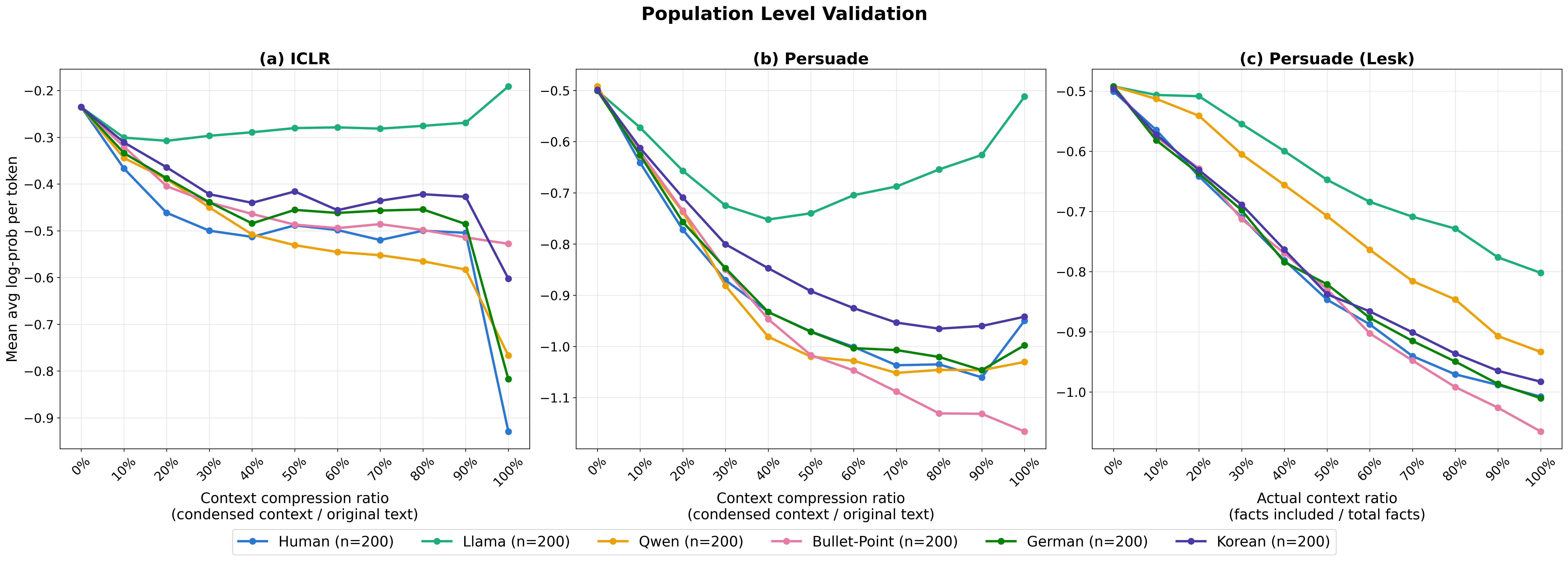}
    \caption{Corpus-level log-probability curves for the ICLR and Persuade datasets. (a) Corpus-level log-probability curve for the ICLR Reviews Dataset using a simple identity extractor, condensation selector. (b) Corpus-level log-probability curve for the Persuade 2.0 essay Dataset using a simple identity extractor, condensation selector. (c) Corpus-level log-probability curve for the Persuade 2.0 essay Dataset using an Equalized Schema-Constrained Fact Extractor with WordNet Canonicalization.}
    \label{fig:iclrpersuade}
\end{figure}

Figure \ref{fig:iclrpersuade}b shows the logprobability curves for the Persuade 2.0 essay dataset. The extractor $E$ here is the simple condensation-based extractor described above in Section \ref{sec:condensation}. Again, the reference LLM (Llama) curve stands out against every other curve.

However, when using a better extractor like the one detailed in Appendix \ref{sec:bestextractor}, we are able to strip away stylistic differences to a greater extent and see an ordering that resembles what we saw for the News dataset: the curve for the Qwen corpus lies in the middle of two extremes, one corresponding to the reference LLM (Llama) curve, and the other corresponding to the Human corpus and its information-preserving variants.

\subsection{Procedure Instantiation for Experiments on the All the News Dataset}
\paragraph{Step 0: Generate the replacement-level document.}
\[
D_{\mathrm{ref}}=M(T).
\]

\paragraph{Step 1: Extract facts.}
Draw $m$ paired extraction samples. For each pair $j=1,\dots,m$, extract $N_j$ facts from the reference and the same number from the evaluated document. Let $F_j^D$ and $F_j^{D_{\mathrm{ref}}}$ denote these
full fact sets, and initialize the retained sets as
\[
C_j^D=F_j^D,
\qquad
C_j^{D_{\mathrm{ref}}}=F_j^{D_{\mathrm{ref}}}.
\]
Fact counts may differ across extraction pairs but are equal within each pair. For any retained fact set $C$ from document $X$, generate reconstructions $\hat X^{(\ell)}$ using $R(T,C,|X|)$ and score each by
\[
s(\hat X^{(\ell)})
=
\frac{1}{|\hat X^{(\ell)}|}
\log P_M(\hat X^{(\ell)}\mid T).
\]
For candidate selection, use $n_{\mathrm{search}}$ reconstructions and let $\bar{s}(C)$ denote their average score.

\paragraph{Step 2: Greedily remove facts.}
At target retention $r$, each extraction pair must retain $\lceil rN_j\rceil$ facts per document. Until this count is reached, consider every possible single-fact removal from the evaluated document's
current set $C_j^D$. Select
\[
f_j^*
=
\operatorname*{arg\,max}_{f\in C_j^D}
\bar{s}\!\left(C_j^D\setminus\{f\}\right),
\]
and update $C_j^D\leftarrow C_j^D\setminus\{f_j^*\}$.
Next, consider every possible single-fact removal from the reference's current set. Select
\[
h_j^*
=
\operatorname*{arg\,min}_{h\in C_j^{D_{\mathrm{ref}}}}
\left|
\bar{s}\!\left(C_j^{D_{\mathrm{ref}}}\setminus\{h\}\right)
-
\bar{s}(C_j^D)
\right|,
\]
where $\bar{s}(C_j^D)$ is the selected candidate's mean score from the preceding search. Update $C_j^{D_{\mathrm{ref}}}\leftarrow
C_j^{D_{\mathrm{ref}}}\setminus\{h_j^*\}$. Repeat these paired removals until both sets contain $\lceil rN_j\rceil$ facts.
This corresponds to greedy selection, as detailed in the main text.

\paragraph{Step 3: Compare at the target retention.}
For each retained set, generate $n_{\mathrm{eval}}$ fresh reconstructions, independently of those used to select deletions. Pool their scores across the $m$ extraction pairs to obtain $mn_{\mathrm{eval}}$ scores per document.

Use a one-sided Mann-Whitney U test with the alternative that the evaluated document's scores tend to be lower than the reference's, rejecting at $p<0.05$. If the test does not reject, stop and
record the current target retention as $r_{\mathrm{stop}}$. Otherwise, proceed to the next lower target retention and repeat Steps 2-3.

The first comparison is at $r=\frac{i}{K}=1$, before any facts are removed. If rejection persists through $r=0$, terminate with $r_{\mathrm{stop}}=0$ and record that all facts were removed without non-rejection.

\paragraph{Step 4: Report the score and identified facts.}
Define
\[
\operatorname{VOLM}(D)=1-r_{\mathrm{stop}}.
\]
The score uses the target retention level; the realized retained fraction in extraction pair $j$ is $\lceil r_{\mathrm{stop}}N_j\rceil/N_j$.

For each extraction sample, also return the removed facts $F_j^D\setminus C_j^D$ and the remaining facts $C_j^D$. When the procedure stops on non-rejection, the removed set identifies the facts selected before the test ceased to detect lower reconstruction scores for the evaluated document.

\subsection{Additional Figures: Distribution of VOLM Scores}
\label{sec:volmdistapp}
\begin{figure}[t!]
\centering
\begin{subfigure}[t]{0.49\textwidth}
    \centering
    \includegraphics[width=\linewidth]{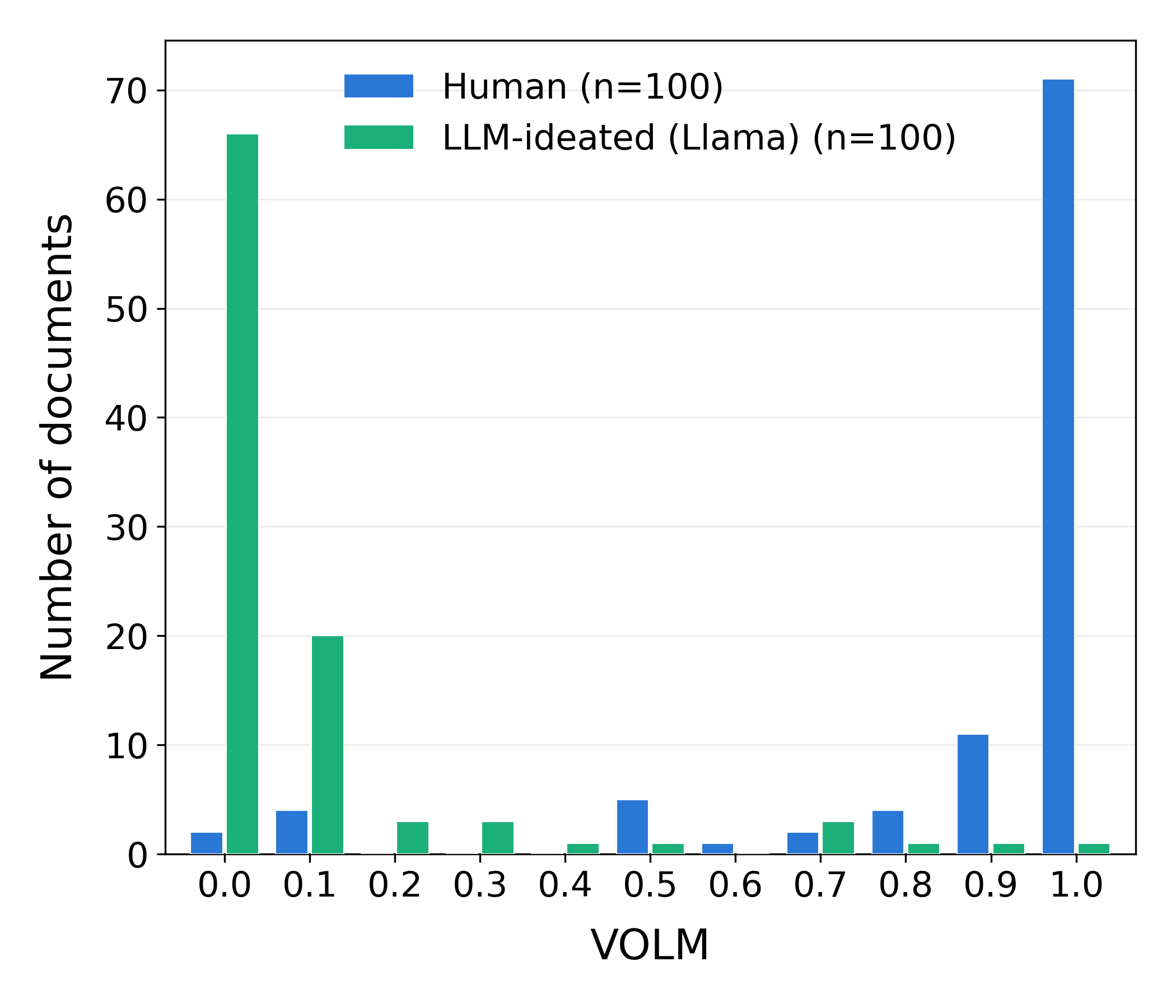}
    \caption{Distribution of VOLM scores for Human-ideated articles from the All the News dataset and LLM-ideated articles sharing the same titles.}
    \label{fig:volmdistappleft}
\end{subfigure}\hfill
\begin{subfigure}[t]{0.49\textwidth}
    \centering
    \includegraphics[width=\linewidth]{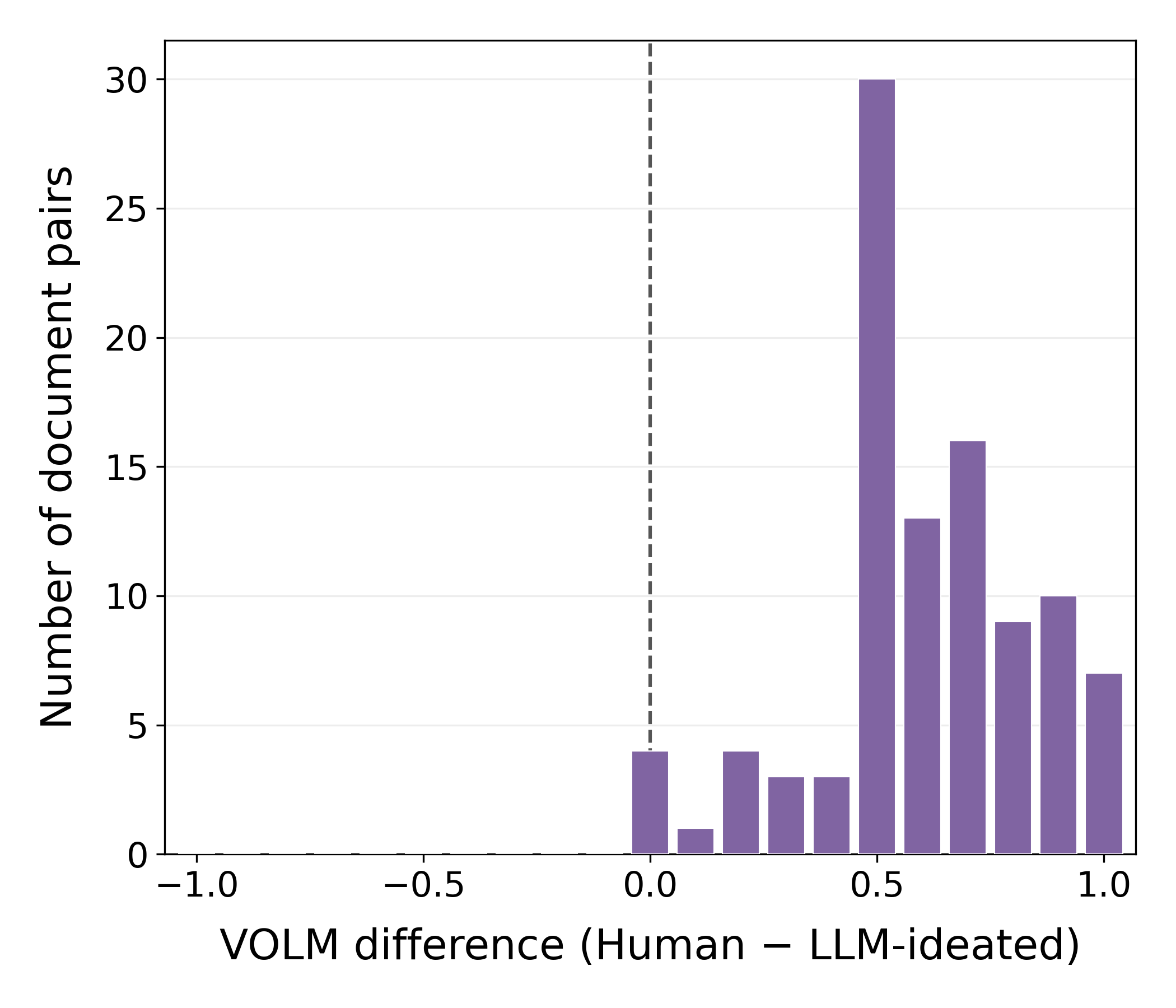}
    \caption{Distribution of VOLM differences between a Human-ideated article and an LLM-ideated article with the same title.}
    \label{fig:volmdistappright}
\end{subfigure}
\caption{Distribution of VOLM scores for human and llm-written articles from the All the News Dataset. The left panel shows the distribution of VOLM scores when using the Preset Selector. The right panel shows the distribution of VOLM(Human)-VOLM(LLM), where articles sharing the same title are paired. We see that the VOLM scores for human articles are always higher (or the same) than those for the corresponding LLM-generated article sharing the same title.}
\label{fig:volmdistapp}
\end{figure}

Figure \ref{fig:volmdistappleft} shows the distribution of VOLM scores for articles from the All the News dataset and a corresponding set of fully LLM-generated articles written using Llama 3.1 8B with the article title and the corresponding human article's length as the only prompt. The Extractor $E$ used was an Equalized Schema-Constrained Fact Extractor with WordNet Canonicalization and the Selector $S$ used was a Preset (ordered) selector. Figure \ref{fig:volmdistappright}shows the distribution of VOLM differences between a human ideated article and it's LLM-ideated counterpart (sharing the same title) for articles from the All the News dataset. The Extractor $E$ used was an Equalized Schema-Constrained Fact Extractor with WordNet Canonicalization and the Selector $S$ used was a Greedy selector.

\subsection{Additional Figures: Population Level VOLM using a Greedy Selector}
\label{sec:populationgreedyapp}
\begin{figure}[t!]
\centering
\includegraphics[width=\linewidth]{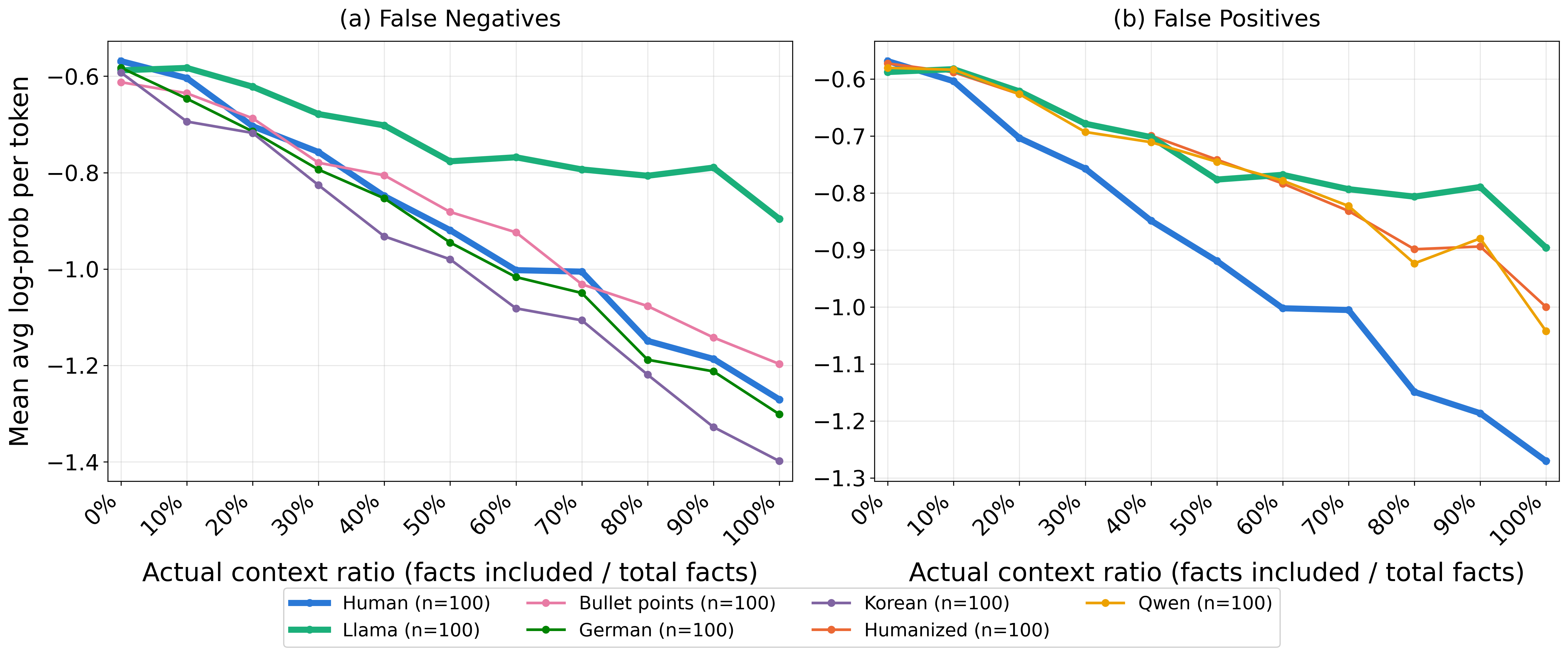}
\caption{Using a Greedy Selector -- Averaged log-probability curves on the All the News dataset. (i) Curves for the reference LLM (Llama)-ideated corpus, human-ideated corpus, and LLM-transformed variants of the human-ideated corpus. (ii) Curves for the reference LLM-ideated corpus, human-ideated corpus, qwen-ideated corpus and humanized corpus.}
\label{fig:popgreedy}
\end{figure}

Figure \ref{fig:popgreedy} shows the population level score curves for every type of corpus we test on. The left panel shows robustness to false negatives -- LLM-transformed variants of human-ideated text should still be classified as high VOLM. The right panel shows robustness to false positives -- humanized versions of LLM-ideated text and text produced by other language models should have low VOLM. On a population level, the human corpus has a VOLM of 1.0, as does the backtranslated German corpus and the bullet-point reconstructed corpus. The backtranslated Korean corpus has a VOLM of 0.9. On the other side, the Qwen and Humanized corpora have VOLM scores of 0.5 and 0.6 respectively.

\subsection{Example of Retained and Removed Facts}
\label{sec:single-doc-facts}

We illustrate the content selected by greedy deletion for the article titled, ``Venezuela detains six military, police officials:
family members, activists.'' This is the same single sample document used in the main text in Figure \ref{fig:two-sides}a, that receives a VOLM score (using a preset selector) of 1.0. We use m=n=8. The document-level greedy procedure stops at 30\% retention, yielding $\operatorname{VOLM}=0.7$. Below, we show extraction sample 0 (seed 1000), which retains two of its ten extracted facts.
The stopping decision uses reconstruction scores pooled across all eight extraction pairs; this sample illustrates one of the resulting fact selections.

\paragraph{Retained facts.}
\begin{itemize}
    \item Venezuelan authorities arrested six members of the country's military and police forces over the weekend.
    \item Nicolas Maduro denied holding political prisoners.
    \item Nicolas Maduro accused the opposition of fomenting violence.
\end{itemize}

\paragraph{Removed facts, in deletion order.}
\begin{enumerate}
    \item Michelle Bachelet called on Maduro on Friday to release prisoners arrested for peacefully protesting.
    \item Rafael Costa was detained in nearby Guarenas on Friday.
    \item Miguel Sisco Mora was arrested in a parking lot in Guatire on Friday afternoon.
    \item Stephanie Sisco demanded on Saturday that the government provide information about her father's whereabouts: ``We demand that the government provide us with information about his whereabouts.''
    \item Nicolas Maduro said he would take the U.N.'s concerns seriously.
    \item Tamara Suju said two high-ranking officials from Venezuela's CICPC forensic police unit were arrested in Guatire.
    \item Penal Forum said two retired Air Force colonels had been arrested in Caracas on Friday afternoon.
\end{enumerate}

These lists describe the selection made along one greedy deletion trajectory. The removed facts represent the information that the model found most surprising (high VOLM). Meanwhile content reconstructed from the retained facts alone did not sustain a detectable difference from the reference at the stopping checkpoint. Observationally, we can see that the removed facts are more informative and specific than the retained ones, which there are only two of. 

\subsection{Susceptibility to stylistic confounding}
\label{sec:wordsretained}
Figure \ref{fig:comparison} shows the issues trivial extractors can suffer from, leaving them vulnerable to stylistic confounding. In spite of subjecting the original document to two LLM operations (selection and reconstruction), the choice of words used in the original document persists, leading to a difference in style between an LLM-generated document and its humanized version. Better extractors can fix this to a large extent, as we show in the main text.
\newcommand{\llamaword}[1]{\colorbox{blue!25}{#1}}
\newcommand{\humword}[1]{\colorbox{orange!35}{#1}}

\begin{figure}[t]
\centering
\footnotesize
\renewcommand{\arraystretch}{1.3}
\resizebox{\textwidth}{!}{%
\begin{tabular}{@{}l >{\raggedright\arraybackslash}p{4.1cm} >{\raggedright\arraybackslash}p{4.1cm} >{\raggedright\arraybackslash}p{4.1cm}@{}}
\toprule
 & \textbf{Original Article} & \textbf{80\% Condensed} & \textbf{Reconstructed} \\
\midrule
\textbf{Llama} &
With its \llamaword{melancholic} melody and \llamaword{heartfelt} lyrics, the song has become a staple of Christmas playlists, providing a sense of validation for those who are struggling to find joy during this time. &
With its \llamaword{melancholic} melody and \llamaword{heartfelt} lyrics, the song has become a staple of Christmas playlists, providing a sense of validation for those who are struggling to find joy during this time. &
With its \llamaword{melancholic} melody and \llamaword{heartfelt} lyrics, the song has become a staple of Christmas playlists, providing a sense of validation for those who are struggling to find joy during this time. \\
\addlinespace[6pt]
\textbf{Humanized} &
With its \humword{mournful} melody and \humword{sincere} lyrics, it has become a standard on Christmastime radio, and it speaks to people who are struggling to find joy at this time of year. &
With its \humword{mournful} melody and \humword{sincere} lyrics, it has become a standard on Christmastime radio, speaking to people who are struggling to find joy at this time of year. &
Its \humword{mournful} melody and \humword{sincere} lyrics have become a staple on Christmastime radio, offering a sense of solidarity and understanding to those who are struggling to find joy. \\
\bottomrule
\end{tabular}%
}
\caption{Persistence of a document's specific word choices across condensation and reconstruction, for a reference Llama-generated article (top) versus the same article after DIPPER humanization (bottom). \llamaword{Blue} marks the Llama article's descriptive word choice, \humword{orange} marks the humanized article's. both pairs persist unchanged through the extractor and reconstructor, even though the two articles start from different wording.}
\label{fig:comparison}
\end{figure}

\subsection{Sample Document}
\label{sec:sampledocument}
\begin{figure}[t!]
    \centering
    \includegraphics[width=0.8\textwidth]{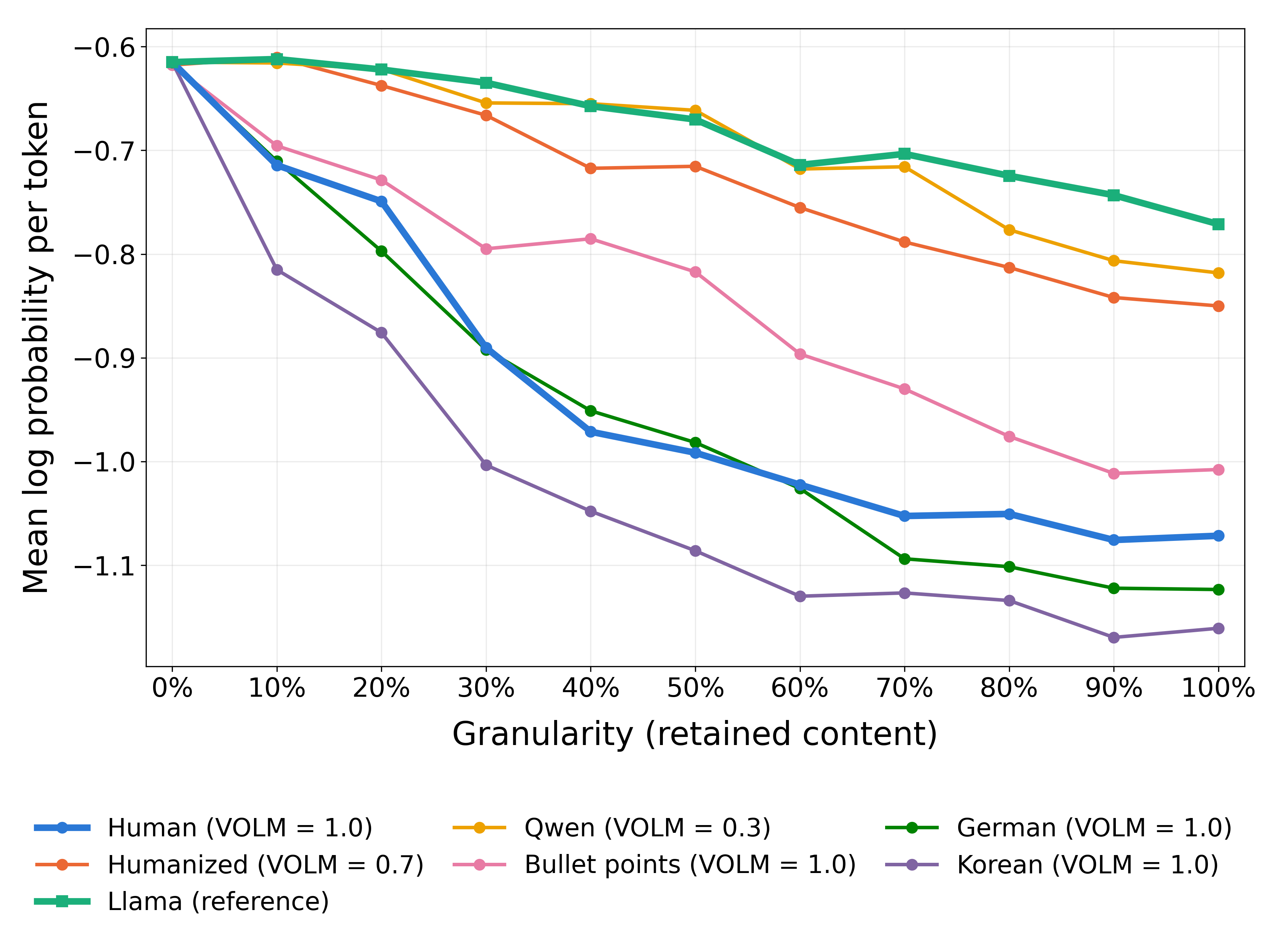}
    \caption{VOLM framework applied to a sample document and multiple LLM-transformed variants.}
    \label{fig:volmsinglefull}
\end{figure}

Figure \ref{fig:volmsinglefull} shows the curves resulting from applying VOLM's complete per-document procedure to a single document, and all its counterparts built to test robustness to false negatives and false positives.

\subsection{Prompts used for Extraction and Reconstruction}

\subsubsection{All the News}

\paragraph{Fact Extraction Prompt (WordNet+Lesk Extractor)}

System prompt (the \texttt{exactly \{target\_count\} facts} line is appended only when fact-count matching to another source is enabled):

\begin{verbatim}
You extract structured facts from a news article as a JSON array of
records. Each record has exactly these fields:

- "subject": the entity performing or central to the fact -- a proper
  name or specific noun phrase exactly as named in the article, never
  a pronoun.
- "action": the single verb that best describes the relation or event,
  in its base/infinitive form (e.g. "invest", "accuse", "say" -- not
  "invested", "to accuse", "is saying").
- "object": the rest of the claim. If the source names multiple
  specific entities (people, places, organizations, products),
  preserve all of their names verbatim here -- never summarize them
  into a vague group like "several companies" or "various officials".
- "quantity": a number, percentage, or amount in this fact, written
  exactly as in the source (e.g. "40%
  null if none.
- "date": a date, year, or time reference for this fact, or null if
  none.
- "quote": if this fact is a direct quotation, the exact quoted words
  verbatim -- never paraphrase a quote -- or null if none.

Rules:
- One record per distinct (subject, action, object) triple. Do not
  merge distinct triples into one record, and do not split a single
  triple into two.
- Preserve every specific named entity, number, and date exactly as
  written in the source -- never generalize, round, or invent one.
- List records in the order the underlying facts first appear in the
  article.
- Output a JSON array of these record objects and nothing else -- no
  preamble, no commentary.
[- You must extract exactly {target_count} facts -- no more, no fewer.]
\end{verbatim}

One-shot example (user turn, then assistant turn) precedes the target document, which is then sent as a final user turn:

\begin{verbatim}
[USER]
Ford Motor Company announced plans in 2020 to invest $1.5 billion in
its plant in Chongqing to produce the Ford Mondeo. The Camp Fire swept
through the town of Paradise, California, in 2018, burning down nearly
19,000 structures and forcing the evacuation of over 50,000 people.
Seattle is home to a burgeoning start-up scene with companies like
Zillow, Expedia, and Tableau Software. "It's a great way to humanize
politicians," said Emily Chen, a 25-year-old social media influencer.

[ASSISTANT]
[{"subject": "Ford Motor Company", "action": "invest", "object": "in
its plant in Chongqing to produce the Ford Mondeo", "quantity": "$1.5
billion", "date": "2020", "quote": null}, {"subject": "Camp Fire",
"action": "destroy", "object": "structures in Paradise, California",
"quantity": "19,000", "date": "2018", "quote": null}, {"subject":
"Camp Fire", "action": "evacuate", "object": "residents of Paradise,
California", "quantity": "50,000", "date": "2018", "quote": null},
{"subject": "Seattle", "action": "house", "object": "a burgeoning
start-up scene with companies including Zillow, Expedia, and Tableau
Software", "quantity": null, "date": null, "quote": null},
{"subject": "Emily Chen", "action": "say", "object": "a 25-year-old
social media influencer, on memes about politicians", "quantity":
null, "date": null, "quote": "It's a great way to humanize
politicians"}]

[USER]
{document_text}
\end{verbatim}

Verb sense disambiguation is \emph{not} an LLM prompt: the extracted
\texttt{action} lemma is canonicalized to a WordNet synset via the
deterministic Simplified Lesk algorithm (word overlap between the
surrounding subject/object context and each candidate sense's gloss),
with no model call involved.

\paragraph{Reconstruction Prompt }

System prompt:

\begin{verbatim}
Write a news article with the following title:
\end{verbatim}

User prompt (generation, context bracketed since it is omitted at the
0\% level and during scoring):

\begin{verbatim}
{title}

Length requirement: write approximately {target_words} words. Your
article must be between {lo_words} and {hi_words} words (inclusive).
Do not mention the word count in the article itself.
[
Additional context:
{context}]
\end{verbatim}

Scoring (teacher-forcing $P(\hat D \mid \cdot)$) always uses the
no-context form of this same user prompt, regardless of the compression
ratio the article was generated at.

\subsubsection{Persuade}

\paragraph{Condensation Prompt (Identity Extractor with a condensation based selector)}

System prompt:

\begin{verbatim}
Condense the provided text to approximately {target_words} words. Your
response MUST be at least {min_words} words. Output only the
condensed text -- no preamble, no word count, no commentary.
\end{verbatim}

User prompt: \texttt{\{original\_essay\_text\}}

\paragraph{Reconstruction Prompt }

System prompt:

\begin{verbatim}
Write a persuasive essay responding to the following prompt:
\end{verbatim}

User prompt:

\begin{verbatim}
Prompt: {assignment}

Length requirement: write approximately {target_words} words. Your
essay must be between {lo_words} and {hi_words} words (inclusive). Do
not mention the word count in the essay itself.
[
Additional context:
{context}]
\end{verbatim}

Scoring again always uses the no-context form.

\subsubsection{ICLR}

\paragraph{Condensation Prompt (Identity Extractor, Condensation-based Selector)}

Identical template to Persuade's, applied to the review text instead of an essay:

\begin{verbatim}
Condense the provided text to approximately {target_words} words. Your
response MUST be at least {min_words} words. Output only the
condensed text -- no preamble, no word count, no commentary.
\end{verbatim}

User prompt: \texttt{\{original\_review\_text\}}

\paragraph{Reconstruction Prompt }

Single user turn (no separate system message):

\begin{verbatim}
You are an expert reviewer for ICLR 2023, a top machine learning
conference.

A research paper is provided below in full text. Read it carefully
and write a thorough peer review following the guidelines below.
{iclr_2023_reviewer_guidelines}

Write your review in free-form prose, structured with the following
sections in order:

**Summary of the paper**
[Summarize what the paper claims to contribute.]

**Strengths and Weaknesses**
[Comprehensive list of strong and weak points.]

**Clarity, Quality, Novelty and Reproducibility**
[Assess the paper's clarity, technical quality, novelty, and
reproducibility.]

**Summary of the review**
[Brief overall summary and recommendation.]

**Rating:** [a single integer from the scale below]
**Confidence:** [a single integer from the scale below]

Rating scale: {iclr_rating_scale}

Confidence scale: {iclr_confidence_scale}

Length requirement: write approximately {target_words} words. Your
review must be between {lo_words} and {hi_words} words (inclusive). Do
not mention the word count in the review itself.
[
Additional context:
{context}]

---

## Paper

{paper_text}
\end{verbatim}

where \texttt{\{iclr\_2023\_reviewer\_guidelines\}} is the official
ICLR 2023 Reviewer Guidelines
(\url{https://iclr.cc/Conferences/2023/ReviewerGuide}), and the
rating/confidence scales are:

\begin{verbatim}
Rating scale:
- 1: Strong Reject
- 3: Reject, not good enough
- 5: marginally below the acceptance threshold
- 6: marginally above the acceptance threshold
- 8: Accept, Good paper
- 10: strong accept, should be highlighted at the conference

Confidence scale (reviewer's certainty about assessment):
- 1: You are not confident in your assessment and your review should
     be taken lightly.
- 2: You are willing to defend your assessment, but it is quite likely
     that you did not understand the central parts of the submission
     or that you are unfamiliar with some pieces of related work.
     Math/other details were not carefully checked.
- 3: You are fairly confident in your assessment. It is possible that
     you did not understand some parts of the submission or that you
     are unfamiliar with some pieces of related work. Math/other
     details were not carefully checked.
- 4: You are confident in your assessment, but not absolutely certain.
     It is unlikely, but not impossible, that you did not understand
     the central parts of the submission or that you are unfamiliar
     with some pieces of related work.
- 5: You are absolutely certain about your assessment. You are very
     familiar with the related work and checked the math/other
     details carefully.
\end{verbatim}

Scoring again always uses the no-context form of the reconstruction
prompt.

\end{document}